\pdfoutput=1

\documentclass[letterpaper]{nature}

\PassOptionsToPackage{dvipsnames,table,xcdraw}{xcolor}

\usepackage[utf8]{inputenc}
\usepackage{color}
\usepackage{xurl}
\usepackage{verbatim}
\usepackage{xspace}
\usepackage{graphicx}
\usepackage{amsmath}
\usepackage{amssymb}
\usepackage{booktabs}
\usepackage{times}
\usepackage[left=1in,right=1in,bottom=1.1in,top=1in]{geometry}
\usepackage{enumitem}
\usepackage{xcolor}
\usepackage{longtable}
\usepackage{makecell}
\usepackage{changepage}
\usepackage{bm}
\usepackage{listings}

\usepackage{array}
\usepackage{tabularx}
\usepackage{adjustbox}
\usepackage{ragged2e}
\usepackage{colortbl}

\usepackage[framemethod=tikz]{mdframed}
\mdfdefinestyle{promptsllm}{
	innertopmargin=1.2\baselineskip,
	innerbottommargin=0.8\baselineskip,
	roundcorner=5pt,
	nobreak,
	singleextra={%
		\draw(P-|O)node[xshift=1em,anchor=west,fill=Yellow!15,draw,rounded corners=5pt]{%
		  \promptVanillaTitle};
	},
}
\newenvironment{promptsllm}[1][Prompts for Evaluating LLMs]{
	\bigskip
	\gdef\promptVanillaTitle{#1}
	\begin{mdframed}[style=promptsllm, backgroundcolor=white]
}{
	\end{mdframed}
	\medskip
}

\usepackage[colorlinks,citecolor=blue,urlcolor=blue]{hyperref}
\usepackage[labelfont=bf, font=footnotesize]{caption}

\hypersetup{
  pdftitle={ToolUniverse: An open platform for democratizing AI scientists},
  pdfauthor={Shanghua Gao, Richard Zhu, Pengwei Sui, Zhenglun Kong, Sufian Aldogom,
             Yepeng Huang, Ayush Noori, Reza Shamji, Krishna Parvataneni,
             Theodoros Tsiligkaridis, Marinka Zitnik},
}

\graphicspath{{./FIG/}}
\newcommand{\xhdr}[1]{\vspace{1mm}\noindent{{\bf #1}}}
\newcommand{\toun}{\textsc{ToolUniverse}\xspace}
\newcommand{\ie}{\emph{i.e.}\xspace}

\newcolumntype{P}[1]{>{\RaggedRight\arraybackslash}p{#1}}
\newcolumntype{Y}{>{\RaggedRight\arraybackslash}X}
\newcolumntype{Z}[1]{>{\hsize=#1\hsize\RaggedRight\arraybackslash\hspace{0pt}}X}
\newcolumntype{L}[1]{>{\raggedright\let\newline\\\arraybackslash\hspace{0pt}}m{#1}}
\newcolumntype{H}{>{\setbox0=\hbox\bgroup}c<{\egroup}@{}}
\definecolor{Gray}{gray}{0.85}
\newcolumntype{b}{>{\columncolor{blue!20}}l}
\newcolumntype{a}{>{\columncolor{Gray}}l}

\title{\begin{center}
ToolUniverse: An open platform for \\ democratizing AI scientists
\end{center}\vspace{-10mm}}

\author
{\small\begin{center}
Shanghua Gao$^{1}$, Richard Zhu$^{1,2,*}$, Pengwei Sui$^{1,*}$, Zhenglun Kong$^{1,*}$, Sufian Aldogom$^{1,*}$, Yepeng Huang$^{1}$, Ayush Noori$^{1}$, Reza Shamji$^{1,2}$, Krishna Parvataneni$^{3}$,  Theodoros Tsiligkaridis$^{4}$, Marinka Zitnik$^{1,5,6,7,\ddag}$ \\[2mm]
{$^{1}$Department of Biomedical Informatics, Harvard Medical School, Boston, MA} \\
{$^{2}$Harvard College, Harvard University, Cambridge, MA} \\
{$^{3}$Massachusetts Institute of Technology, Cambridge, MA} \\
{$^{4}$MIT Lincoln Laboratory, Lexington, MA} \\
{$^{5}$Kempner Institute for the Study of Natural and Artificial Intelligence, Harvard University, Cambridge, MA} \\
{$^{6}$Broad Institute of MIT and Harvard, Cambridge, MA} \\
{$^{7}$Harvard Data Science Initiative, Cambridge, MA} \\
{$^*$Co-second authors}; {$\ddag$Correspondence: marinka@hms.harvard.edu}\\
\end{center}
}

\begin{document}

\maketitle

\spacing{1.15}

AI scientists are AI agents powered by conversational large language models that carry out complex analyses by calling scientific tools, querying datasets, and interacting with experimental systems, other agents, and human researchers~\cite{gao2024agent}. Despite rapid progress, most are bespoke and difficult to reuse because they lack standardized interfaces for tools and data~\cite{boiko2023autonomous,swanson2025virtual}. Scientific research requires that AI scientists act on biological and physical systems rather than reason through text alone, operating within an environment of tools, agentic skills, replayable agentic memory, and new agentic capabilities introduced in the fast advancing field~\cite{wang2023scientific,gao2024agent}. The tools, skills, and capabilities supporting this interaction vary in purpose and complexity, require distinct inputs and parameters, depend on specific runtime environments, and run locally or through application programming interfaces (APIs) and laboratory systems. The tools among them span modalities from protein and nucleic acid sequences and molecular structures to microscopy images and biological networks, and their overlapping functionality complicates selecting and executing an appropriate tool for an objective expressed in natural language.

We present \toun, a platform for building AI scientists across open- and closed-weight language models and agents. \toun implements scientific tool use through six modules, Tool Manager, Tool Finder, Tool Caller, Tool Composer, Tool Discoverer, and Tool Optimizer, that together enable AI scientists to discover, retrieve, execute, and compose tools from a library of more than 2,700 scientific tools as of July 2026 (Figure~\ref{fig:fig1}a; Supplementary Note~\ref{app:overview_toun} and~\ref{app:tool-categories}). \toun is available at \url{https://aiscientist.tools} and accessible through a Python API, HTTP, the Model Context Protocol (MCP), and direct function calls (Supplementary Note~\ref{sec:build_ai_scientists_examples}).

\toun defines a standard for AI-tool interaction that each tool implements so that AI scientists can call it. A tool specifies its intended use in natural language, a schema naming and typing all inputs and outputs with their constraints, and a backend-agnostic invocation format, so the same call invoked in different contexts can execute a Python library, a machine learning model, a dataset endpoint, or a laboratory instrument without tool-specific configuration (Supplementary Note~\ref{app:core-components}). The composition of the library by tool type and its coverage by data modality and application area are summarized in Supplementary Tables~\ref{table:tool_example} and~\ref{table:tool_modality}. The library's tools are also accessible through over 130 research skills (Supplementary Table~\ref{table:skill_categories}).

\toun manages tools through six components that cover how they are registered, found, called, composed, created, and refined (Figure~\ref{fig:fig1}b--g; Supplementary Note~\ref{app:core-components}). Tool Manager registers local functions and remote services behind a single interface, so tools with different backends are invoked the same way; Tool Finder retrieves the most relevant tools for a natural language request by combining keyword, language model, and embedding based search; and Tool Caller executes the selected tool with validated inputs, returning structured results or actionable errors through Python, the \toun API, MCP, and HTTP. Tool Composer chains tools into sequential, parallel, or feedback driven workflows; Tool Discoverer creates new runnable tools from natural language descriptions and automatically validates them, with human review required before registration; and Tool Optimizer refines tool descriptions to match observed behavior. We benchmark Tool Finder across its keyword, embedding, and language model search strategies, and evaluate the embedding strategy under a range of interchangeable open and hosted encoders; guided by these results, we add strong performing encoders as built-in, selectable options. We also benchmark \toun on standard scientific tasks, where an agent or language model equipped with \toun outperforms both the same agent without it and a purpose-built biomedical agent~\cite{huang2025biomni} (Supplementary Note~\ref{app:module-benchmark}).

Language models and agents invoke \toun in context through tool specifications or through agentic frameworks such as Gemini CLI and Claude Code; end-to-end examples are in Supplementary Note~\ref{sec:build_ai_scientists_examples}.
Building on these interfaces, \toun can be used in two complementary ways. Researchers developing new models can leverage its integrated environment of tools and agentic skills to teach models tool use through reinforcement learning, where agents acquire policies for selecting and parameterizing tools and improve over long agentic runs, as in ATHENA~\cite{gao2026ai}. Existing agents and language models can integrate \toun to expand their available tools through standardized retrieval and invocation interfaces, as in Virtual Lab~\cite{swanson2025virtual}, Medea~\cite{medea2026}, and GeneAgent~\cite{wang2025geneagent}.

Beyond individual tools, \toun's research skills are natural language workflows that orchestrate multiple tools into complex analyses. A biologist describes a goal in plain language, for example performing differential expression analysis of RNA-seq data, assessing a candidate drug target, or reviewing disease evidence, and the matching skill runs the end-to-end workflow. These skills span genomics, transcriptomics, proteomics, variant interpretation, drug discovery, and single cell and spatial biology (Supplementary Note~\ref{app:skills}), with examples provided in Supplementary Note~\ref{app:case-studies-realworld} for cancer target assessment, chemical compound analysis against a druggable target, and clinical trial analysis. \toun also verifies, logs, and governs tools across their lifecycle, confirming through test cases and review of input-output traces that each tool functions as described before use, recording tool calls and arguments through optional logging hooks for inspection and audit, and applying additional governance safeguards to mitigate safety risks~\cite{tang2025risks} (Supplementary Note~\ref{app:tool-evaluation}).

We envision \toun growing as an open platform. Because every tool implements the same AI-tool interaction standard, \toun scales without changes to the core platform, whether contributors add local tools, register remote tools over MCP, or generate them with Tool Discoverer (Supplementary Note~\ref{app:core-components}). Community-contributed tools and skills are incorporated only if they meet explicit criteria, namely conformance to the \toun protocol, test cases covering typical and edge conditions, human expert review of input-output traces, and tool quality thresholds (Supplementary Note~\ref{app:tool-evaluation}). As human-AI co-science evolves further, reproducibility and safety grow in importance. \toun supports reproducibility through versioned open-source tools and audit logging, and safety through human-in-the-loop oversight of consequential actions~\cite{tang2025risks}.

\clearpage

\spacing{1}

\xhdr{Data availability.}
A complete, versioned list of the sources accessed by each tool is provided with the code at \url{https://github.com/mims-harvard/ToolUniverse}. Supplementary Information reports case studies and benchmarking obtained from live tool executions. All data supporting the findings of this study are available within the paper and its Supplementary Information and from the public repositories and web services accessed by its tools. \toun is available at \url{https://aiscientist.tools}, with documentation at \url{https://github.com/mims-harvard/ToolUniverse} and additional information at \url{https://zitniklab.hms.harvard.edu/ToolUniverse}.

\xhdr{Code availability.}
\toun source code is available under Apache License~2.0 at \url{https://github.com/mims-harvard/ToolUniverse}. Development follows semantic versioning with tagged, numbered releases, and the version described in this manuscript is v1.3.1. A citable archive of this release is deposited on Zenodo~\cite{tooluniverse_zenodo}. \toun can be installed from the Python Package Index with \texttt{pip install tooluniverse} (\url{https://pypi.org/project/tooluniverse}), and the \toun Model Context Protocol (MCP) connector is at \url{https://aiscientist.tools/mcp}. Documentation, tutorials, and contribution guidelines are provided in the repository.

\xhdr{Acknowledgments.}
We thank Nicholas Yang and Yuchang Su for their contributions to \toun. We gratefully acknowledge the support of NSF CAREER 2339524, ARPA-H Biomedical Data Fabric (BDF) Toolbox Program, Harvard Data Science Initiative, Amazon Faculty Research, Google Research Scholar Program, AstraZeneca Research, Roche Alliance with Distinguished Scientists (ROADS) Program, Sanofi iDEA-iTECH Award, GlaxoSmithKline Award, Boeh\-ringer Ingelheim Award, Merck Award, Optum AI Research Collaboration Award, Pfizer Research, Gates Foundation (INV-079038), Chan Zuckerberg Initiative, John and Virginia Kaneb Fellowship at Harvard Medical School, Biswas Computational Biology Initiative in partnership with the Milken Institute, Harvard Medical School Dean's Innovation Fund for the Use of Artificial Intelligence, and the Kempner Institute for the Study of Natural and Artificial Intelligence at Harvard University. Any opinions, findings, conclusions, or recommendations expressed in this material are those of the authors and do not necessarily reflect the views of the funders.

\xhdr{Competing interests.}
None.

\clearpage

{\spacing{1}

\begin{figure}[ht]
\centering
\includegraphics[width=1.0\textwidth]{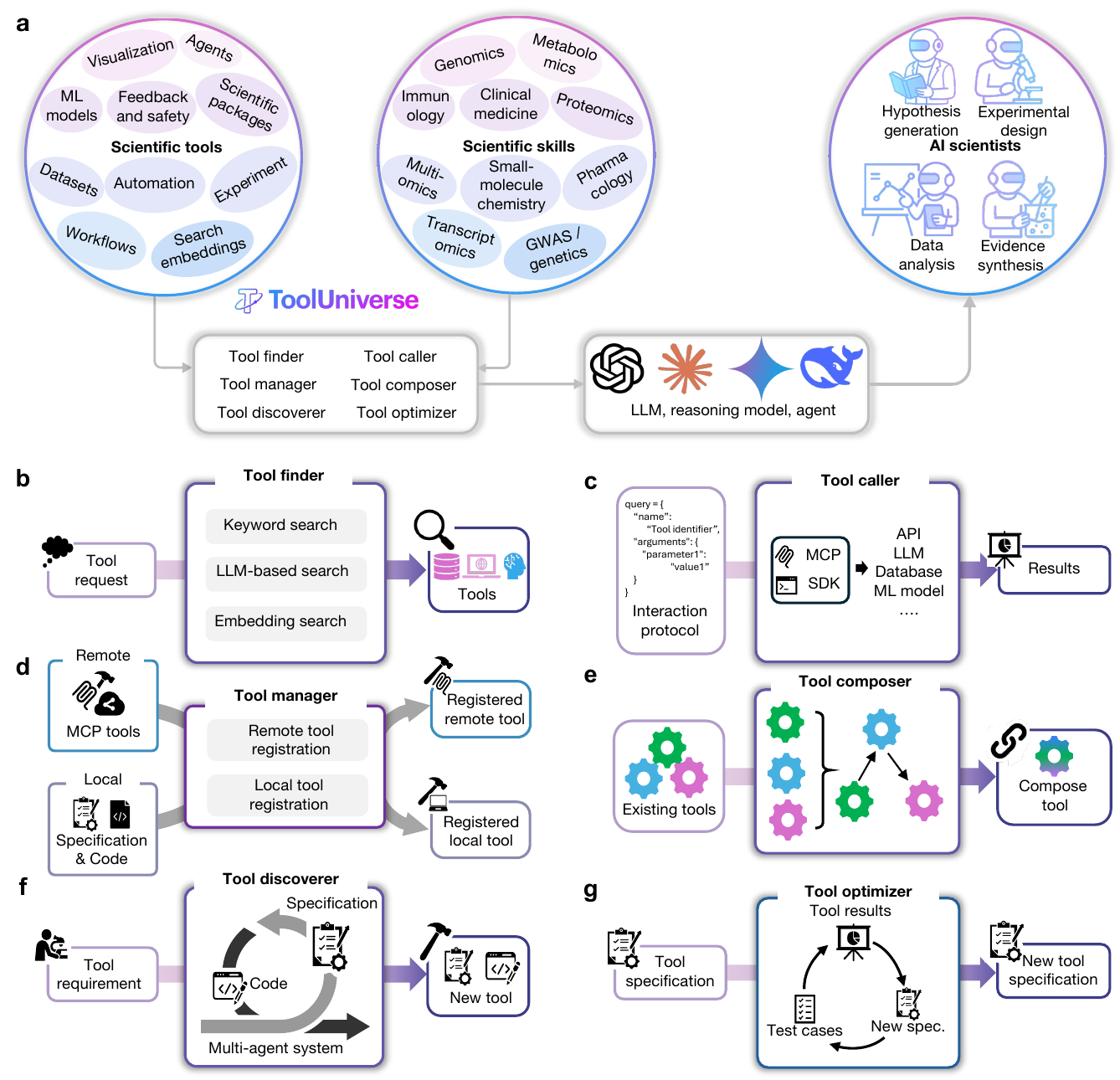}
\caption{}
\label{fig:fig1}
\end{figure}
\clearpage

\xhdr{Figure~\ref{fig:fig1}:}
\textbf{a.} \toun is an open platform for building AI scientists that integrates 2{,}700 scientific tools and 130 research skills as of July 2026. The tools span ML models, agents, datasets, workflows, scientific packages, laboratory automation and experimentation, visualization, search embeddings, and human feedback and safety (left; full taxonomy in Supplementary Table~\ref{table:tool_type}), and are exposed through the research skills grouped by scientific area, including genomics, transcriptomics, proteomics, pharmacology, small molecule chemistry, clinical medicine, immunology, metabolomics, multi-omics, and GWAS and genetics (middle; Supplementary Table~\ref{table:skill_categories}). Six modules (Tool Finder, Tool Caller, Tool Manager, Tool Composer, Tool Discoverer, and Tool Optimizer; detailed in panels b--g) connect the platform, through the AI-tool interaction standard, to large language models, reasoning models, and agents (GPT, Claude, Gemini, and DeepSeek). Connected to \toun, these models act as AI scientists that perform research tasks, including hypothesis generation, experimental design, data analysis, and evidence synthesis (right). End-to-end examples of building AI scientists on \toun are provided in Supplementary Note~\ref{sec:build_ai_scientists_examples}, and a quantitative evaluation of the platform in Supplementary Notes~\ref{app:module-benchmark} and~\ref{app:case-studies-realworld}.
\textbf{b.} Tool Finder identifies relevant tools using keyword search, LLM-based in-context search, and embedding-based similarity search.
\textbf{c.} Tool Caller validates inputs, dynamically loads tools, dispatches calls through \toun.run() or MCP, and returns tool outputs.
\textbf{d.} Tool Manager integrates local and remote tools, registering local tools via JSON descriptions and decorators and adding remote tools through MCP for privacy- or dependency-constrained setups.
\textbf{e.} Tool Composer chains multiple tools into composite workflows, supporting sequential, parallel, and feedback-driven orchestration of heterogeneous tools.
\textbf{f.} Tool Discoverer is a multi-agent system that generates new tools from natural-language requirements, performing definition synthesis, automated code generation, validation, and iterative refinement to produce runnable tools.
\textbf{g.} Tool Optimizer is a multi-agent system that iteratively refines tool descriptions to improve clarity, accuracy, and usability, combining test generation, execution analysis, and feedback-driven refinement.
ML, Machine Learning; LLM, Large Language Model; MCP, Model Context Protocol; API, Application Programming Interface; SDK, Software Development Kit.

\clearpage

}

\clearpage

\section*{References}

{
\bibliographystyle{naturemag}
\bibliography{refs}
}

\clearpage

\setcounter{section}{0}
\setcounter{figure}{0}
\setcounter{table}{0}
\setcounter{equation}{0}
\setcounter{page}{1}
\renewcommand{\thesection}{S\arabic{section}}
\renewcommand{\thefigure}{S\arabic{figure}}
\renewcommand{\thetable}{S\arabic{table}}
\renewcommand{\thepage}{S\arabic{page}}
\renewcommand{\theequation}{\arabic{equation}}

\renewcommand{\xhdr}[1]{\vspace{1.7mm}\noindent{{\bf #1.}}}
\captionsetup{font=normalsize}
\setlength{\tabcolsep}{3pt}

\thispagestyle{empty}
\spacing{1}

\begin{center}
{\Large Supplementary Information for\\[3mm]
{\bf ToolUniverse: An open platform for democratizing AI scientists}\\[3mm]}
Shanghua Gao$^{1}$, Richard Zhu$^{1,2,*}$, Pengwei Sui$^{1,*}$, Zhenglun Kong$^{1,*}$, Sufian Aldogom$^{1,*}$, \\ Yepeng Huang$^{1}$, Ayush Noori$^{1}$, Reza Shamji$^{1,2}$, Krishna Parvataneni$^{3}$, \\ Theodoros Tsiligkaridis$^{4}$, Marinka Zitnik$^{1,5,6,7,\ddag}$ \\[2mm]
{$^{1}$Department of Biomedical Informatics, Harvard Medical School, Boston, MA} \\
{$^{2}$Harvard College, Harvard University, Cambridge, MA} \\
{$^{3}$Massachusetts Institute of Technology, Cambridge, MA} \\
{$^{4}$MIT Lincoln Laboratory, Lexington, MA} \\
{$^{5}$Kempner Institute for the Study of Natural and Artificial Intelligence, Harvard University, Cambridge, MA} \\
{$^{6}$Broad Institute of MIT and Harvard, Cambridge, MA} \\
{$^{7}$Harvard Data Science Initiative, Cambridge, MA} \\{$^*$Co-second authors}; {$\ddag$Correspondence: marinka@hms.harvard.edu}\\[2mm]
\toun website is at \url{https://aiscientist.tools}\\
\toun code is at \url{https://github.com/mims-harvard/ToolUniverse}\\
\toun package is at \url{https://pypi.org/project/tooluniverse}
\end{center}

\vspace{1cm}

{\spacing{1}
\noindent This PDF file includes:
\begin{description}[labelsep=2em, align=left, itemsep=0em]
\item[] Supplementary Notes 1 to 9
\item[] Supplementary Figures 1 to 9
\item[] Supplementary Tables 1 to 4
\end{description}
\vspace{1cm}
}

\clearpage

{\spacing{1}

\section{Supplementary Note: Overview of \toun}\label{app:overview_toun}

\toun is an ecosystem that exposes tools to any open or closed AI model, \ie, large language model (LLM), agent, or large reasoning model (LRM), allowing users to build AI research assistants (\ie, AI scientists) without additional training or finetuning. To achieve that, \toun provides the user-specified LLM/LRM/agent with a scientific toolkit (Supplementary Table~\ref{table:tool_type}). While advanced models possess planning and reasoning capabilities, scientific research cannot be conducted through reasoning alone. \toun addresses this by exposing executable tools that return experimental data, database query results, and model predictions. \toun defines an AI-tool interaction standard that makes tools understandable to LLMs, LRMs, and AI agents through natural-language descriptions and typed schemas, abstracting away backend differences. 

\toun is extensible and allows tools to be easily added, optimized, or created. It hosts a toolkit of over 2,700 scientific tools and is designed for integration with language models, agents, and reasoning models. Current tools in \toun include: foundation models, finetuned LLMs, LRMs, and other ML models exposed as callable endpoints; agentic planners and tool routers; domain libraries and simulators; and systems for human-in-the-loop feedback and lab automation with instrument control. \toun also provides data and retrieval utilities, such as data sources, knowledge bases, vector search with embedding generators, and complete retrieval-augmented generation (RAG) pipelines. For integration and governance, the ecosystem offers external service connectors, typed API clients, privacy guardrails, safety checklists, compliance controls, and audit logs. \toun supports high-level scientific and operational workflows through visualization dashboards, experiment design tools with ELN/LIMS integration, and workflow engines and orchestration schedulers. Despite the backend heterogeneity of these tools, which span machine learning models, AI agents, software utilities, robotics, databases, and APIs, all are presented to the AI scientist through the AI-tool interaction standard, which we describe next.

\xhdr{AI-tool Interaction Standard}
\toun defines an AI-tool interaction standard that makes the backend agnostic to users and simplifies the addition of new tools. Each tool includes three elements: (i) a natural-language description of the tool's purpose and expected behavior, (ii) a typed schema listing every input parameter and output field with names, types, and constraints, and (iii) a uniform invocation format that abstracts away backend differences. By supplying this information within an AI model's context window, \toun enables any LLM, reasoning model, or agent to understand and invoke tools without tool-specific configuration.

\xhdr{Core Components}
This standard supports six core modules. Tool Manager registers local functions and remote services, Tool Finder searches for options from over 2,700 candidates based on user requirements, and Tool Caller executes tools with validated arguments and structured returns. For complex tasks, Tool Composer assembles multiple tools into workflows. Tool Discoverer generates new tools from natural language descriptions, and Tool Optimizer refines tool descriptions through iterative testing and feedback.
These components and the AI-tool interaction standard enable creating AI scientists: specialized AI models that combine reasoning with access to tools to perform multi-step research tasks.

\section{Supplementary Note: AI-tool Interaction Standard in \toun}\label{app:tool-interaction-protocol}
\toun is designed to support an ecosystem of tools with a wide range of abilities. This note specifies the AI-tool interaction standard through which every one of them is managed and presented to the client in the same form.

\subsection{Tool Description Schema}
Supplementary Figure~\ref{fig:tool_spec} shows the tool description schema in \toun.
The AI-tool interaction standard exposes every tool via a description containing its name; a functional description; a list of parameters, where each parameter is explicitly defined with its own name, description, data type, and required status; and a return schema that shows the data structure of the returned data.
An example tool description is shown in Supplementary Figure~\ref{fig:tool_example}.
This description is provided to clients such as LLMs, reasoning models, AI agents, and human users to help them understand how to use the tool.
For instance, when the client is an LLM, the tool description is supplied within its context window, thereby granting it the necessary information to interact with tools from \toun.

\subsection{Tool Interaction Schema}
\toun processes requests through a standard interaction schema, illustrated in Supplementary Figure~\ref{fig:tool_interact}.
All interactions are formatted as a single string that encodes a function call, specifying the desired tool's name and its input arguments.
This schema provides the foundation for all interactions with \toun, enabling tool search, calling, discovery, optimization, and composition.

\subsection{Accessing Tools from \toun} Using the AI-tool interaction standard, \toun provides an interface for human users and AI agents to access tools.
The tool can be invoked by executing:
\begin{verbatim}
tooluniverse.run(tool_call_schema)
\end{verbatim}
where \verb|tool_call_schema| is a dictionary following the interaction schema: \{`name': name of the tool, `arguments': parameters required by the tool\}.
This approach abstracts away backend complexity. Regardless of a tool's underlying implementation, it is presented to the client (the user of the tool, such as LLMs, reasoning models, AI agents) through the AI-tool interaction standard.
To use a tool, the client consults the tool description to construct a request that adheres to the interaction schema. This request is then sent to \toun via either a local interface or a remote http connection. Tool Caller processes the request, executes the specified tool, and returns the results to the client.
This eliminates the need for complex configurations, regardless of backend or runtime differences. For example, users can query a new database without writing database-specific SQL, run machine learning models without configuring GPUs or environments, or access web-based lab equipment through a single, standardized tool call request.

\section{Supplementary Note: Core Modules of \toun}\label{app:core-components}

\toun organizes the ecosystem into six modules for tool management. Tool Manager integrates tools into \toun through local and remote registration. Tool Finder searches for suitable options from over 2,700 candidates based on user requirements. Tool Caller executes tools with validated arguments and structured returns. For complex tasks, Tool Composer assembles multiple tools into workflows. Tool Discoverer generates new tools from natural language descriptions. Tool Optimizer refines tool descriptions through iterative testing and feedback.

\subsection{Tool Manager}\label{app:tool-manager}
Tool Manager is designed to simplify the process of adding new tools to \toun through two registration modes.
For local tools, which require no special dependencies, only a JSON description (including name, descriptions, typed arguments, and configurations) and a corresponding callable function are needed. Once registered, any connected AI model can invoke them.
For remote tools, which may have specialized dependencies or restricted access, \toun integrates them through MCP by registering the server address, which loads all hosted tools into \toun's tool list. After registration, both local and remote tools expose the same AI-tool interaction standard, making backend differences invisible to AI models.

\xhdr{Local Tool Registration}
For local tool registration, to add a tool to \toun, both a tool configuration and a corresponding tool class are required. The tool configuration is a dictionary that specifies the tool according to the AI-tool interaction standard and includes the necessary settings for its execution. The tool class defines an initialization function that sets up the tool based on its description, as well as a run function that processes tool call arguments in accordance with the interaction schema. Local tool registration within the Tool Manager is facilitated through an easy-to-use decorator function, \texttt{register\_tool(Class\_name, tool\_config)}, which decorates the tool class as illustrated in Supplementary Figure~\ref{fig:tool_local_reg}. Here, \texttt{Class\_name} is the string name of the tool class, and \texttt{tool\_config} is the configuration containing both the tool description and required settings. Once registered, the tool is automatically integrated into \toun without further manual configuration. This registration process allows users to incorporate custom tools into \toun, enabling coordination with other existing tools and empowering the creation of customized AI scientists.

\xhdr{Remote Tool Registration}
Remote Tool Registration enables the integration of tools that are private, require specialized configurations, or operate within restricted environments, and therefore cannot be made publicly available.
Once registered remotely, these tools are added to \toun and can be accessed and executed in the same manner as standard tools.

To achieve this, the Tool Manager includes an MCP Auto Loader Tool that accepts the address of an MCP server and registers all of its tools into the \toun's tool list. After the MCP Auto Loader Tool has loaded the remote tools, they are integrated into the \toun with the same functionality as other tools.
For the remote side, \toun supports two methods for setting up a remote tool. The first is building standard tools that support MCP. To further simplify the process and make remote tool registration identical to local registration, the Tool Manager also provides a decorator function, \texttt{register\_\allowbreak{}remote\_\allowbreak{}tool(Class\_\allowbreak{}name, tool\_\allowbreak{}config, mcp\_\allowbreak{}config)}, which decorates the tool class as illustrated in Supplementary Figure~\ref{fig:tool_remote_reg}. In this function, the \texttt{Class\_name} and \texttt{tool\_config} parameters are the same as those used in \texttt{register\_tool(Class\_name, tool\_config)}, while \texttt{mcp\_config} defines the configuration for the MCP server, such as the host address and port used for the service.

\subsection{Tool Finder}\label{app:tool-finder}
The Tool Finder retrieves tools from a library of more than 2,700 scientific tools using three strategies, each trading off speed, semantic understanding, and compute cost. The input is a natural language query from the client, describing the task they wish to achieve or the specific capabilities required from the tools. The three strategies are: keyword search using TF-IDF scoring, LLM-based in-context search that reasons about multi-step or abstract requests, and embedding search that encodes queries and tool descriptions using a fine-tuned GTE-Qwen2-1.5B model.

\xhdr{Keyword search} Keyword search parses a user's query through a multi-stage workflow involving tokenization via regular expressions, the removal of over 45 common English stop words, and suffix-based stemming using 20 morphological rules to reduce words to their root form. To capture multi-word concepts, this method also generates n-grams (bigrams and trigrams). These processed keywords and phrases are then matched against a pre-built index of tool descriptions, which has undergone the same processing. 
Relevance is scored using a term frequency-inverse document frequency (TF-IDF) algorithm, calculated as:
$\text{Relevance} = \text{TF} \times \text{IDF} \times \log(1 + \text{QueryFrequency})$,
where \text{TF} measures how often a term appears in a document, \text{IDF} reflects how unique the term is across all documents, and \text{QueryFrequency} indicates how often the term appears in the user's query.
The scoring model adjusts relevance scoring by applying a hierarchical bonus structure to the relevance score. Matches found in a tool's name receive the largest weight with a 2.0× bonus multiplier, followed by a 1.5× multiplier for exact phrase matches within descriptions. This keyword search approach provides a keyword-based search solution that operates independently of machine learning models and can be used across different resource levels.

\xhdr{LLM in-context search}
The LLM in-context search uses the reasoning capabilities of a Large Language Model to interpret user intent more holistically. Rather than relying on simple keyword matching, a detailed prompt for tool selection is constructed. This prompt contextualizes the user's task description with the tool descriptions of a candidate set of tools in \toun. The LLM is then tasked with analyzing this rich context to infer a suitable tool or sequence of tools required to fulfill the user's request. This method is used for interpreting complex, multi-step, or abstract queries that demand logical inference. While its application can be constrained by the finite context window of the model, it offers flexibility in understanding abstract goals.
The LLM in-context search is powered by the agentic tool implementation in \toun. This allows the user to provide a configuration file that defines the prompt and tool descriptions, without needing to manage the backend LLM inference processes.

\xhdr{Embedding search}
Embedding search is a vector-based method that retrieves tools by matching the semantic similarity between a user's query and the tool's description. To achieve this, we finetune the GTE-Qwen2-1.5B language embedding model using pairs of synthetic user queries and augmented tool descriptions, training it to understand the connection between a user's intent and a tool's function.
The process involves two stages. First, in an offline indexing stage, each tool's description is passed through the embedding model to generate a semantic vector that captures its meaning. These vectors are then stored and indexed in a specialized vector database. Later, during the online querying stage, a user's natural language query is converted into a query vector using the same model. Finally, relevant tools are discovered by calculating the cosine similarity between the user's query vector and all the tool vectors in the database, identifying the closest matches.

\subsection{Tool Caller}\label{app:tool-caller}
The Tool Caller is the primary execution engine in \toun and is accessible through four interfaces: direct Python function calls, the \toun Python API, MCP, and HTTP API, making tools accessible from any programming environment or remote client. Upon initialization, the Tool Caller is configured with a manifest of available tools, including their descriptions and settings. To mitigate the significant system overhead associated with loading all tools simultaneously, it employs a dynamic loading strategy. A specific tool is loaded into memory only upon its first request and is then cached for a duration to efficiently handle subsequent calls. During this loading process, the Tool Caller injects the necessary configurations, such as API endpoints and authentication keys, into the corresponding tool class.

When a tool execution request is received, the Tool Caller first parses it to extract the tool name and arguments. It then performs a rigorous validation check, ensuring the provided arguments conform to the data types and structural requirements defined with the AI-tool interaction standard. Once validated, the Tool Caller dispatches the arguments to the tool's primary execution method, such as \texttt{run()}, as illustrated in Supplementary Figure~\ref{fig:run_tool_example}. The resulting output is then returned to the client through \toun's communication protocols. If any step in this process fails, from loading to validation or execution, the system generates and returns a descriptive error message. This feedback mechanism helps the client diagnose the issue and revise the request accordingly.

\subsection{Tool Composer}\label{app:tool-composer}
For complex tasks, users can create new composite tools by combining existing ones. Tool Composer chains tools with heterogeneous backends into multi-step workflows. Using the Tool Caller for direct in-code execution, Tool Composer generates a container function that exposes both the Tool Caller and \toun as in-line, executable primitives.
The container function, implemented as \texttt{compose(arguments, tooluniverse, call\_tool)}, serves as the execution backbone for Tool Composer. It contains the logic for coordinating different types of tools so they work together in a single workflow. The \texttt{arguments} parameter specifies the tool call arguments that follow the interaction schema of \toun, the \texttt{tooluniverse} is an instance of \toun that provides all available functions that \toun can support, and the \texttt{call\_tool} parameter is a callable interface of Tool Caller that abstracts the invocation of individual tools in \toun.
By integrating these components, the container function enables three execution patterns: (i) chaining the output of one tool into the input of the next, (ii) broadcasting a single query across multiple tools in parallel, and (iii) constructing agentic loops in which a reasoning model generates tool calls, executes them, and incorporates the feedback for adaptive multi-step analysis. As illustrated in Supplementary Figure~\ref{fig:tool_compose}, a composed tool can run several literature search tools concurrently and then invoke a summarization agent to synthesize the findings, demonstrating heterogeneous workflow construction in which each step is driven by tool execution.

\subsection{Tool Discoverer}\label{app:tool-discoverer}
Tool Discoverer automatically generates new tools, including both tool definitions and executable code, from high-level natural language descriptions. Leveraging the workflow composition capabilities of \toun via Tool Composer, we construct an agentic multi-stage workflow that transforms a plain-text functional request into a runnable tool with limited human intervention. The process integrates tool discovery, structured definition generation, code implementation, and iterative quality refinement, ensuring that the final tool adheres to functional requirements and ecosystem conventions. Here, ``limited human intervention'' refers to the automated generation loop and does not bypass human review: every generated tool, like all tools in \toun, undergoes the human-in-the-loop expert review described in the Supplementary Note on Evaluation of Tools (Supplementary Note~\ref{app:tool-evaluation}) before it is added to the platform.

Through iterative feedback loops combining search, analysis, and code optimization, the system progressively refines the generated tool until it reaches predefined quality criteria. The process terminates when target scores are achieved or iteration limits are reached, producing tools intended for downstream use.

\xhdr{Core principles and architecture}
The Tool Discoverer system is built around four core principles: pattern-guided generation that reuses functional patterns and conventions from existing tools to ensure ecosystem consistency; structured definition synthesis that transforms unstructured requests into tool definitions; automated code generation and validation that produces executable implementations with integrated testing; and iterative refinement that uses feedback from analysis and testing to drive targeted improvements.

The system integrates three specialized components working in sequence. The SpecificationGenerator converts the natural language description and retrieved reference implementations into a structured definition, including tool name, description, parameter definitions, return schema, and category metadata. The ImplementationGenerator produces code that follows standard software engineering practices, includes complete dependency handling, integrates with the \toun registry, and implements error handling for failure cases. The QualityEvaluator assesses the generated tool across functionality, reliability, maintainability, performance, and test coverage, scoring each dimension from 0--10 and computing an overall weighted score.

\xhdr{Generation process}
The generation workflow begins with the discovery stage, which accepts a tool description as inputs. It searches GitHub, PyPI, and existing \toun tools for reference implementations using multiple search strategies (semantic similarity and keyword-based search). Results are compiled into a final set of references.
Next, the definition stage uses an agentic tool to produce a complete tool configuration. This includes tool name, descriptions, parameter definitions with type annotations and descriptions, JSON-schema-based return type definition, and other metadata. The output conforms to \toun schema requirements and naming conventions, enabling immediate downstream integration.
The implementation phase employs template-driven code generation to produce runnable code, incorporating required imports, type-hinted function signatures, error handling, integration hooks via the @register\_tool decorator in \toun, and a module structure suitable for integration.
Finally, the system conducts multiple quality assessments, combining code analysis, dynamic testing with auto-generated test cases, and execution profiling. Each dimension is scored, with a target minimum of 9/10 for deployment-oriented settings. An iterative refinement loop applies targeted improvements until the threshold is met.
Generated tools are packaged for immediate inclusion in \toun, producing a JSON configuration file with metadata, a source file with the complete implementation, and dependency information for reproducible installation.

\subsection{Tool Optimizer}\label{app:tool-optimizer}
Tool Optimizer is designed to refine tool descriptions with respect to clarity, accuracy, and interpretability for models.
Using Tool Composer, we build an agentic tool description optimization workflow where tool descriptions are optimized through an iterative multi-round process combining automated test case generation, real tool execution analysis, and feedback refinement.
Through feedback-driven iterations, the system refines both tool descriptions and parameter descriptions, eliminating redundancy between them while checking alignment through empirical validation. The optimization process automatically terminates when quality thresholds are met or maximum iterations are reached, producing descriptions that aim to support tool use for AI agents.

\xhdr{Core principles and tools}
The Tool Optimizer employs a multi-round iterative optimization strategy to automatically enhance tool documentation quality. The optimizer is built on three core principles: (1) test-driven optimization that validates descriptions against actual tool execution results, (2) multi-dimensional quality assessment across six standardized criteria, and (3) feedback-driven improvement that uses insights from previous optimization rounds to guide subsequent iterations.
The optimizer implements a compositional architecture consisting of three specialized components working in concert. The TestCaseGenerator creates diverse test scenarios based on tool configurations and adaptively generates targeted test cases in later rounds using feedback from previous iterations. The DescriptionAnalyzer examines the alignment between existing descriptions and actual tool behavior by analyzing test execution results, then generates optimized descriptions for both tool-level and parameter-level documentation that better reflect the tool's true functionality while ensuring consistency with real usage patterns and eliminating redundancy between tool and parameter descriptions. Finally, the DescriptionQualityEvaluator provides objective scoring on a 0-10 scale across six quality dimensions: clarity, accuracy, completeness, conciseness, user-friendliness, and redundancy avoidance.

\xhdr{Optimization process}
The optimization process begins with initial test case generation followed by tool execution to gather execution results. The system then enters an iterative optimization loop where each round generates enhanced test cases based on previous feedback, analyzes accumulated test results to propose improved descriptions, optimizes both tool and parameter descriptions, and evaluates quality against predefined thresholds. The process continues until either the satisfaction threshold is met (typically 8.0/10 for deployment-oriented settings) or the maximum iteration limit is reached (default: 3 rounds). This adaptive approach ensures comprehensive coverage while maintaining runtime constraints.

\section{Supplementary Note: Building AI scientists with \toun}
\label{sec:build_ai_scientists_examples}
A customized AI scientist can be developed by integrating \toun with LLMs, reasoning models, and AI agents. In this configuration, the LLMs and reasoning models provide the core capabilities for reasoning and tool usage, while \toun serves as the scientific environment for interaction and experimentation. The development process involves three steps: (i) installing \toun with a single command (\texttt{pip install tooluniverse}), (ii) connecting \toun to the chosen model so it can access the tools provided by \toun, and (iii) instructing the model to use these tools to address a given scientific problem.

Once the setup is complete, the AI scientist operates as follows: given a user instruction or task, it formulates a plan or hypothesis, employs Tool Finder in \toun to identify relevant tools, and iteratively applies these tools to gather information, conduct experiments, verify hypotheses, and request human feedback when necessary. For each required tool call, the AI scientist generates arguments that conform to the AI-tool interaction standard, after which Tool Caller executes the tool and returns the results for further reasoning.

The models used to construct AI scientists can include LLMs, reasoning models, and AI agents. LLMs may be API-based, such as Claude or GPT, or open-weight models, such as LLaMA, DeepSeek, or Qwen. Large reasoning models enhance problem-solving capabilities by applying built-in chains of thought to analyze the current step before interacting with \toun. Agentic systems, such as Gemini CLI or Claude Code, integrate reasoning models with agentic feedback loops to autonomously manage multi-step problem solving and tool use. In addition to general-purpose agents, \toun can be paired with specialized agents designed for specific scientific domains by integrating domain-specific tools.
The following sections present three examples of building AI scientists: (i) LLM in-context use, (ii) MCP agent use, and (iii) specialized agents.

\begin{figure}[ht]
\centering
\includegraphics[width=1.0\textwidth]{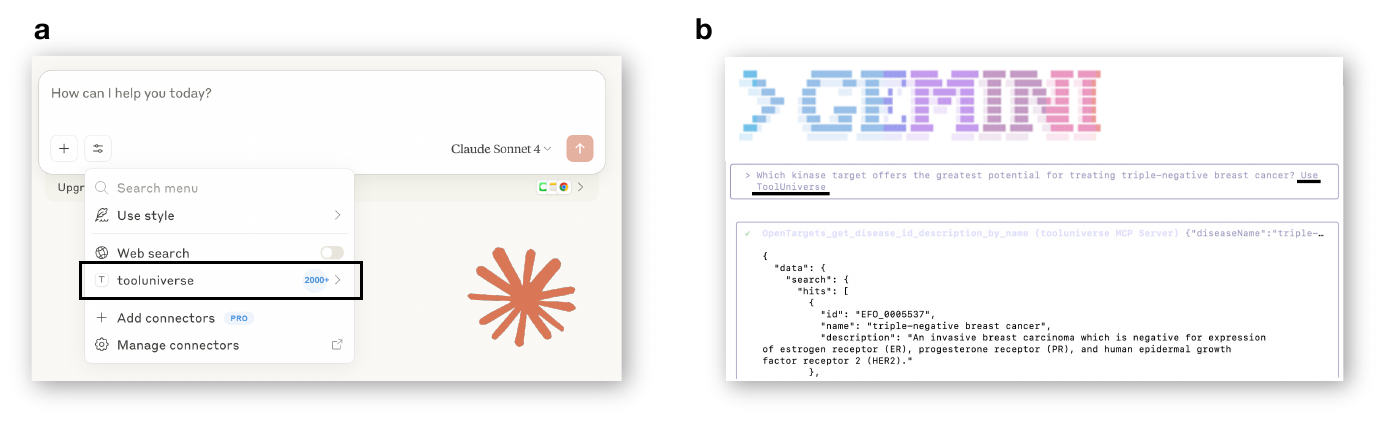}
\caption{Building AI scientists on \toun. (a)~Using \toun with a large language model in-context: the Claude desktop application registers \toun as a Model Context Protocol (MCP) connector, exposing its tools (here, more than 2{,}000) for selection and use directly within a chat session. (b)~Using \toun with an AI agent: the Gemini CLI connects to \toun over MCP and autonomously calls tools to answer a research query (here, prioritizing kinase targets for triple-negative breast cancer), showing an issued tool call to a \toun MCP tool and its returned result.}
\label{fig:si_claude_gemini}
\end{figure}

\subsection{Building an AI Scientist from an LLM or an LRM}
We illustrate building an AI scientist from an LLM, such as Claude, together with \toun (Supplementary Figure~\ref{fig:si_claude_gemini}a). The process involves three steps.

\begin{enumerate}
  \item Install \toun with a single command and install the Claude desktop app.
  
  \item Open the Claude Desktop and navigate to 
  \texttt{Settings → Developer → Edit Config}.  
  Set up the configurations as follows:
  
  \lstset{
      basicstyle=\ttfamily\footnotesize,
      numbers=none,
      xleftmargin=0pt,
      frame=none
  }
  \begin{lstlisting}[language=bash]
  {
      "mcpServers": {
        "tooluniverse": {
          "command": "uv",
          "args": [
              "--directory",
              "path_to_ToolUniverse/src/tooluniverse",
              "run",
              "tooluniverse-mcp-claude"
          ]
        }
      }
  }
  \end{lstlisting}
  
  \item Launch the Claude Desktop, select tools in \toun for the desired tasks.
\end{enumerate}

\subsection{Building an AI Scientist from an AI Agent}
We demonstrate how to build an AI scientist using AI agents, such as the Gemini CLI, together with \toun (Supplementary Figure~\ref{fig:si_claude_gemini}b).
While AI agents can automatically leverage Tool Finder to identify the tools required for specific tasks, the process of creating an AI scientist involves two steps.

\begin{enumerate}
  \item Install \toun with a single command and install the Gemini CLI.
  
  \item Open the setting configuration file for Gemini CLI.  
  Set up the configurations as follows:
  
  \lstset{
      basicstyle=\ttfamily\footnotesize,
      numbers=none,
      xleftmargin=0pt,
      frame=none
  }
  \begin{lstlisting}[language=bash]
{
  "mcpServers": {
    "tooluniverse": {
      "command": "uv",
      "args": [
        "--directory",
        "/path/to/your/gemini_running_env",
        "run",
        "tooluniverse-smcp-stdio"
      ]
    }
  },
}
  \end{lstlisting}
\end{enumerate}

\subsection{Building an AI Scientist from a Specialized AI Agent}

Specialized AI agents are trained on specific types of tasks, allowing them to become experts in particular domains, such as TxAgent~\cite{gao2025txagent} for precision therapeutics, GeneAgent~\cite{wang2025geneagent} for gene-set analysis, and SpatialAgent~\cite{spatialagent} for spatial biology.
\toun can be used not only by these specialized AI agents during inference but also as a real-world environment for agent training. For example, in TxAgent~\cite{gao2025txagent}, \toun serves as the scientific environment in which TxAgent can interact. Through reinforcement learning, TxAgent learns how to use tools within \toun and manage complex therapeutic tasks. 

\section{Supplementary Note: Tools in \toun}\label{app:tool-categories}

\toun is an ecosystem that contains over 2,700 scientific tools covering essential scientific research domains. \toun provides built-in tool categories covering machine learning models, AI agents, software utilities, expert feedback systems, robotics, databases, embedding stores, data archives, and APIs, each serving specific computational and analytical requirements.

\xhdr{Agentic tools}
Agentic tools operate autonomously to perform complex tasks using LLMs.
Each agent is configurable with custom prompts and tool descriptions, supporting multiple backend models, including ChatGPT and Gemini.
\toun includes agentic tools for literature summarization, code analysis, hypothesis generation, experiment planning, and results analysis.
By defining the prompts and tool descriptions in the configuration file, one can quickly build an agentic tool.

\xhdr{Scientific software package tools}
To support scientific coding, scientific package tools provide information about Python-based scientific computing libraries such as NumPy, Pandas, and SciPy. These tools offer installation instructions, usage examples, and documentation links, implementing dual-source data retrieval from PyPI APIs with local backup information.

\xhdr{Database tools}
Database tools manage structured scientific resources such as DrugBank vocabulary datasets, clinical trial records, and molecular databases. They support integrations with tabular, hierarchical, XML-based, and graph-structured data. These tools provide capabilities for text-based search, field-level filtering, configurable result limits, and metadata return schemas. Built-in search, filtering, and indexing features can be reused when incorporating new databases.

\xhdr{API integration tools}
API integration tools communicate with external scientific data sources using standard protocols such as RESTful APIs or GraphQL. Through these tools, users can access resources like FDA drug databases, OpenTargets disease–target associations, PubChem compound information, and many other databases, all with structured error handling and response validation. New tools can be incorporated by updating the API server URL, provided the common protocol is maintained.

\xhdr{Expert feedback tools}
Expert feedback tools connect AI scientists with human experts, allowing AI scientists to request human suggestions or approval whenever necessary.
This tool includes a server that connects the system with human experts, along with a user interface through which experts can provide responses.
When a user calls the expert feedback tool, the request is redirected by the tool caller in \toun to a server, which forwards it to the human expert interface. Human experts can receive the request and provide their own insights and judgments. Their feedback is then sent back through the server as a tool response to the user.
This approach supports consultation with human experts for complex scientific decisions and interpretations, combining automated analysis with expert validation.

\xhdr{Machine learning tools}
Machine learning tools apply predictive and generative models to scientific use cases, such as disease–target scoring~\cite{opentargets2024}, disease-state prediction, gene-gene interaction~\cite{transcriptformer, Li2024}, gene dependency analysis~\cite{DepMap2024}, ADMET prediction~\cite{swanson2024admet}, binding affinity prediction~\cite{passaro2025boltz}, and beyond. Since running environments for machine learning models often require specialized setups and hardware (e.g., GPUs), which can be difficult to deploy, \toun uses a remote registration scheme. This approach allows models to run on private servers while still being exposed as tools within \toun. New machine learning models can be quickly integrated into \toun through remote tool registration.

\xhdr{Embedding store tools}
Embedding store tools manage vectorized representations of scientific data. Scientific data is first transformed into embeddings using embedding models and then stored in a database. \toun employs FAISS to enable semantic search, similarity matching, and data retrieval over these embedding databases.

\section{Supplementary Note: Pre-built Research Skills}\label{app:skills}
To make \toun directly usable by experimental and computational biologists, the platform provides a library of pre-built research \emph{skills}. Each skill is a natural-language-invokable workflow that orchestrates multiple tools to complete an end-to-end scientific task and return a reproducible result with provenance. As of this writing, \toun includes more than 130 research skills spanning genomics, transcriptomics and gene expression, proteomics, variant interpretation and clinical genetics, drug discovery and pharmacology, single-cell and spatial omics, microbiome and infectious disease, and structural biology. A user states a goal in natural language and the platform routes the request to the appropriate skill, which executes the underlying tools, performs the analysis, and returns results. Representative skills include RNA-seq differential expression with DESeq2 (\texttt{tooluniverse-rnaseq-deseq2}), drug-target validation (\texttt{tooluniverse-drug-target-validation}), comprehensive disease research (\texttt{tooluniverse-disease-research}), clinical variant interpretation with ACMG classification (\texttt{tooluniverse-variant-interpretation}), single-cell RNA-seq analysis (\texttt{tooluniverse-single-cell}), and GWAS-driven drug discovery (\texttt{tooluniverse-gwas-drug-discovery}). Skills follow a common specification and can be contributed and extended by the community. Supplementary Table~\ref{table:skill_categories} groups the available skills by application and data-modality area.

\subsection{Worked examples}\label{app:skills-worked}
To make the skill workflow concrete, we trace two requests end to end---from the natural-language ask, through the tools invoked, to the returned result. All results below are from live tool executions.

\textbf{A single-call skill (sequence retrieval).} A user asks ``get me the canonical TP53 protein sequence.'' The platform routes to \texttt{EnsemblSeq\_\allowbreak{}get\_\allowbreak{}id\_\allowbreak{}sequence}, which resolves the Ensembl protein ID \texttt{ENSP00000269305} and returns the 393-amino-acid sequence (\texttt{MEEPQSDPSV}\ldots), the canonical p53 isoform. This is the simplest skill shape: one natural-language request mapped to one validated tool call.

\textbf{An end-to-end analysis skill (RNA-seq differential expression).} A user asks ``run differential expression on my airway RNA-seq data, dexamethasone-treated versus untreated, controlling for donor,'' providing a count matrix (public benchmark data: airway smooth-muscle cells from four donors, treated and untreated; \cite{himes2014rna}, GEO GSE52778). The skill \texttt{tooluniverse-rnaseq-deseq2} loads the matrix (63{,}677 genes $\times$ 8 samples), builds the design \texttt{\textasciitilde{} cell + dex}, and runs R DESeq2 through the \texttt{run\_\allowbreak{}deseq2\_\allowbreak{}analysis} tool (contrast treated vs untreated, $\alpha=0.05$). It returns 4{,}028 differentially expressed genes at $\mathrm{padj}<0.05$ (2{,}211 up, 1{,}817 down); the most significant genes include the glucocorticoid-responsive \textit{DUSP1} ($\log_2\!FC=2.95$), and \textit{CRISPLD2}---the gene the original study followed up---is also strongly induced ($\log_2\!FC=2.63$, $\mathrm{padj}=4.4\times10^{-46}$). Chaining the induced gene set into the same tool's GO-enrichment step (\texttt{enrichgo}) returns 58 enriched biological-process terms after redundancy reduction; alongside airway smooth-muscle terms, these include \emph{cellular response to glucocorticoid stimulus} (BH-adjusted $p=4.7\times10^{-3}$), recovering the expected mechanism of action. The entire analysis runs from a single natural-language request, with every number traceable to the tool that produced it.

\begin{center}
\small
\begin{adjustbox}{max width=\linewidth}
\begin{tabular}{@{}lll@{}}
\toprule
\textbf{Step} & \textbf{Tool (operation)} & \textbf{Result} \\
\midrule
Sequence retrieval & \texttt{EnsemblSeq\_get\_id\_sequence} & TP53 protein, 393 aa \\
Differential expression & \texttt{run\_deseq2\_analysis} (deseq2) & 4{,}028 DEGs (2{,}211 up, 1{,}817 down) \\
Top DEG & \texttt{run\_deseq2\_analysis} (deseq2) & \textit{CRISPLD2}, $\log_2\!FC=2.63$, padj $=4.4\times10^{-46}$ \\
Pathway enrichment & \texttt{run\_deseq2\_analysis} (enrichgo) & glucocorticoid response, adj.\ $p=4.7\times10^{-3}$ \\
\bottomrule
\end{tabular}
\end{adjustbox}
\end{center}

\section{Supplementary Note: Evaluation of Tools}\label{app:tool-evaluation}
To ensure the correctness and reliability of tools before their inclusion in \toun, we implemented a multi-step evaluation process. This evaluation was designed both to validate tool functionality and to ensure scientific utility.
Each tool passes through the following step-by-step review before it is admitted to \toun:
\begin{enumerate}
    \item \textbf{Automated validation.} For each tool, we generate diverse sample inputs covering typical use cases and edge conditions and record the outputs, and automated optimizer and checker tools test the tool against these samples, verifying that its behavior is consistent with its declared functionality and refining its description for accuracy and usability before expert review.

    \item \textbf{Human-in-the-loop review.} Every tool undergoes human expert review before inclusion, and the form of this review depends on the tool's provenance. (i)~For tools that wrap an authoritative data source (for example national or international resources such as the NIH and FDA), the source is itself authoritative, so the review verifies that the integration in \toun returns the same records as the original source rather than re-validating the underlying data, as for the openFDA tools. (ii)~For tools backed by a peer-reviewed publication, the underlying method has already been validated in the literature, so the review confirms that the wrapped tool reproduces the published behavior on reference inputs, as for the Boltz-2 docking tool. (iii)~For tools maintained by their original developers (for example a model served through the developer's own API), the tool is evaluated jointly with that team to confirm correct integration; this was the case for the ESM Cambrian (ESMC) protein language model, which was incorporated in collaboration with the EvolutionaryScale team. (iv)~For all other tools, scientific experts perform a full review of representative input-output traces, assessing correctness, interpretability, and consistency with expected domain knowledge, as for \toun's in-house scientific-calculation tools (for example enzyme-kinetics and statistical-test utilities). In all cases, this human evaluation provides a safeguard against erroneous or misleading outputs.

    \item \textbf{Decision and admission.} On the basis of the human review and the automated checks, the tool is either approved and registered into \toun, or returned for revision, to the contributor or to the automated generation and optimization loop, and re-reviewed before a final decision.

    \item \textbf{Regular maintenance and bug reporting.} For continued reliability of tools within \toun, we perform regular maintenance and monitoring. A structured bug reporting system allows issues to be identified by users and promptly addressed by the \toun team. These practices provide ongoing quality assurance and safeguard the long-term usability of the ecosystem.
\end{enumerate}

Many tools in \toun are drawn from established, previously validated resources, and the provenance-dependent review above leverages that prior validation to reduce the risk of introducing spurious or unverified functionality into the ecosystem.
These evaluation measures aim to ensure that tools incorporated into \toun meet standards of correctness, reproducibility, and scientific reliability, providing a foundation for AI-assisted discovery.

\clearpage

\section{Supplementary Note: Quantitative Evaluation of \toun on End-to-End Tasks}\label{app:module-benchmark}

This note describes the end-to-end benchmark used to evaluate \toun's higher-level modules (Tool Composer, Tool Discoverer, and Tool Optimizer). Because these modules act by assembling tools into complete workflows rather than as standalone predictors, we assess them through end-to-end task accuracy, the standard on which agentic systems such as Biomni~\cite{huang2025biomni} report.

\xhdr{Benchmarks} We evaluate on two LAB-Bench~\cite{laurent2024labbench} subtasks: DbQA, which poses biomedical database questions that require querying authoritative resources, and SeqQA, which poses nucleotide- and protein-sequence questions with deterministic answers. To make the comparison to Biomni exact, we evaluate on the identical test items released by Biomni (60 DbQA and 70 SeqQA questions) and reuse Biomni's option construction and scoring procedure without modification, so every system is graded on the same questions under the same protocol; our reconstruction reproduces Biomni's released items and gold answers item for item.

\xhdr{Systems and protocol} We compare \toun-enabled agent systems with matched no-\toun agent systems and with Biomni. We run the controlled with-versus-without-\toun ablation in two independent agent settings, one built on Claude Code and one on Codex, so that the effect of adding \toun's tools is measured under two different base models; within each setting the two conditions share an identical agent and base model and differ only in whether the \toun tools are available. In the Claude setting, the \toun-enabled and no-\toun Claude Code agent systems both use Claude Opus~4.8. In the Codex setting (GPT-5.5), each question is sent to a fresh Codex process, with the \toun condition receiving the same tool-use instruction and the no-\toun condition receiving only the benchmark prompt; each Codex condition is run three independent times and reported as mean accuracy $\pm$ sample standard deviation. Biomni values are its reported accuracies on the same items. Answers are parsed and scored with Biomni's procedure (abstention or an unparseable response counts as incorrect).

\xhdr{Results} In a controlled ablation, equipping an agent with \toun's tools substantially improves its end-to-end task accuracy (Figure~\ref{fig:si_labbench_endtoend}). Each \toun-enabled agent is compared against an identical agent and base model that differ only in whether \toun's tools are available, so the gain is attributable to the platform itself. It is large on DbQA, where accuracy rises from 56.7 to 78.3 for the Claude Code agent (a 21.6-point improvement) and from 81.1 to 92.8 for the Codex agent (an 11.7-point improvement); both agents also improve on SeqQA, from an already high baseline, rising from 94.3 to 99.5 for the Claude Code agent (a 5.2-point improvement) and from 91.4 to 96.2 for the Codex agent (a 4.8-point improvement). The effect reproduces across two agents built on different base models, Claude Opus~4.8 through Claude Code and Codex (GPT-5.5), indicating that the benefit is a general property of \toun and does not depend on a particular model or agent framework. Codex accuracies are means of three independent runs whose run-to-run standard deviations are small (at most 3.5 points, and 1.0 point once \toun is added on DbQA), confirming that the improvement is stable across runs.

The \toun-augmented agents also outperform Biomni, a purpose-built biomedical agent, on every subtask, evaluated on Biomni's own released items and scoring. Biomni reaches 74.4 on DbQA and 81.9 on SeqQA; the \toun-augmented Claude Code agent reaches 78.3 and 99.5 and the Codex agent 92.8 $\pm$ 1.0 and 96.2 $\pm$ 1.6, so a general-purpose agent equipped with \toun surpasses a specialized biomedical system on both tasks. The Claude numbers understate what the agent can do: its base model refuses some of the questions on safety grounds, and each refusal is counted as a wrong answer, which lowers the Claude scores whether or not \toun is available. Even under this penalty, the \toun-augmented Claude Code agent still outperforms Biomni on both subtasks.

\begin{figure}[htbp]
\centering
\includegraphics[width=\textwidth]{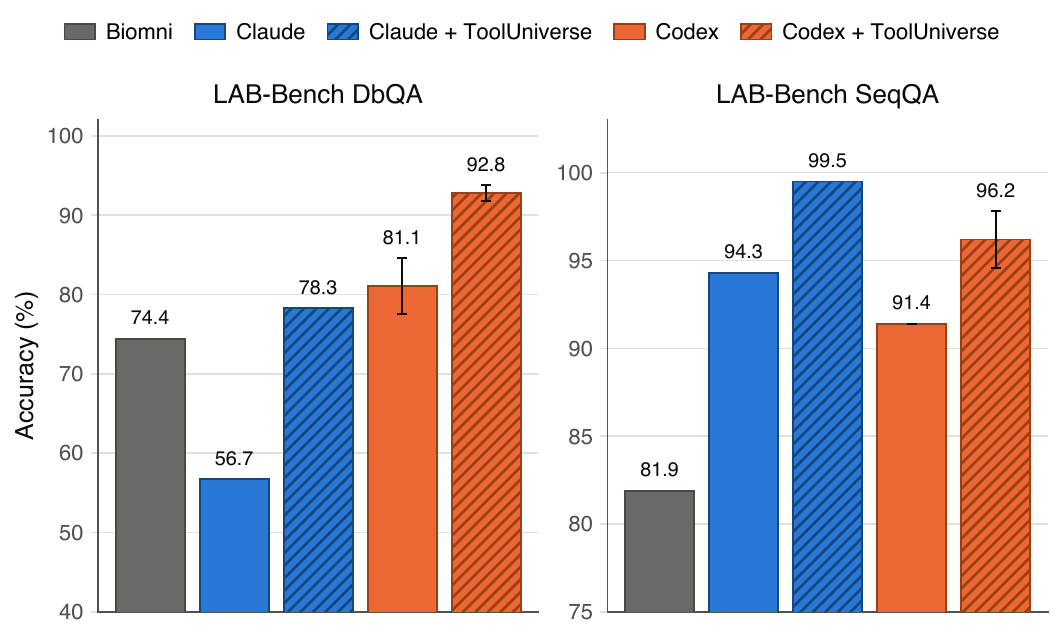}
\caption{End-to-end accuracy (\%) on LAB-Bench DbQA and SeqQA, shown as separate panels. Hatched bars are our \toun-augmented agents; solid bars are the baselines: the identical agent and base model run without \toun, and Biomni (gray). Within each agent, Claude Opus~4.8 (via Claude Code) and Codex (GPT-5.5), adding \toun (the hatched bar, right of its matched baseline) improves accuracy, and both \toun-augmented agents exceed Biomni on both subtasks; all systems are evaluated on the same publicly released items under the same scoring. Codex bars are the mean $\pm$ sample standard deviation over three independent runs (error bars); Claude and Biomni values are point estimates. A subset of questions are declined by the Claude base model (Claude Opus~4.8) under its safety policy and scored as incorrect, which lowers both Claude bars regardless of tool availability. Each panel's vertical axis begins above zero to resolve differences among the high-scoring systems.}
\label{fig:si_labbench_endtoend}
\end{figure}

\clearpage

\section{Supplementary Note: Step-by-step real-world case studies}\label{app:case-studies-realworld}
\sloppy

To illustrate how researchers use \toun in practice, we present three end-to-end case studies in which an AI scientist---an LLM connected to \toun---answers a real research question by autonomously selecting and chaining tools. Each case states the question in natural language, then follows the workflow as it composes tools across databases, predictive models, and analysis services, reporting the concrete result at each step. The workflows go beyond database lookup, integrating human genetics, functional-genomics screens (DepMap CRISPR essentiality~\cite{DepMap2024}), protein structure (AlphaFold~\cite{jumper2021highly} and experimental PDB structures), cheminformatics and experimental bioactivity (ChEMBL~\cite{zdrazil2024chembl}), and machine-learning property prediction (ADMET-AI~\cite{swanson2024admet}). The cases span (i) context-selective oncology-target assessment, (ii) structure- and prediction-guided chemical strategy, and (iii) clinical-trial biostatistics, and all results are from live tool executions. Across the three, the value of the workflow lies less in retrieving evidence than in reconciling evidence that points in different directions, and in refining an initial question as the evidence accrues. Strengths and limitations are summarized at the end.

\subsection{Case 1: Is \textit{BLM} a viable anticancer target, and is its chemical probe developable?}\label{case:blm}
\emph{Research question: ``\textit{BLM} loss-of-function predisposes to cancer by compromising genome maintenance, which raises a safety concern for inhibiting \textit{BLM}. Is \textit{BLM} nevertheless a context-selective anticancer target, and is its inhibitor ML216 a developable lead?''} The AI scientist approaches this in stages, refining the question as the evidence accrues.

\begin{enumerate}
  \item \textbf{Resolve and annotate the gene.} \texttt{OpenTargets\_\allowbreak{}multi\_\allowbreak{}entity\_\allowbreak{}search\_\allowbreak{}by\_\allowbreak{}query\_\allowbreak{}string} (API Tool) maps \textit{BLM} to \texttt{ENSG00000197299} (``BLM RecQ like helicase''), and \texttt{OpenTargets\_\allowbreak{}get\_\allowbreak{}target\_\allowbreak{}gene\_\allowbreak{}ontology\_\allowbreak{}by\_\allowbreak{}ensemblID} (API Tool) annotates a nuclear ATP-dependent 3$'$--5$'$ DNA helicase acting in DNA repair and homologous recombination (GO:0006281, GO:0043138, GO:0000724), with roles in replication-fork processing (GO:0031297) and telomere maintenance (GO:0061820, GO:0090656). A genome-maintenance helicase is the kind of gene whose loss promotes tumorigenesis, so its cancer-predisposition genetics counsel caution about inhibition rather than support it.
  \item \textbf{Quantify the cancer-risk genetics.} \texttt{OpenTargets\_\allowbreak{}get\_\allowbreak{}diseases\_\allowbreak{}phenotypes\_\allowbreak{}by\_\allowbreak{}target\_\allowbreak{}ensembl} (API Tool) returns 464 disease associations. Bloom syndrome leads (Genomics England 0.96; ClinVar/EVA 0.95), followed by ``inherited cancer-predisposing syndrome'' (EVA 0.95) and a broad malignancy spectrum that includes colorectal cancer, acute myeloid and lymphoblastic leukemia, lymphoma, hereditary breast and ovarian cancer, and prostate carcinoma. These associations confirm that \textit{BLM} loss is clinically consequential for genome instability.
  \item \textbf{Classify the patient's variant.} \texttt{ClinVar\_\allowbreak{}search\_\allowbreak{}variants} (API Tool) returns 495 pathogenic \textit{BLM} variants, among them \texttt{c.520C>T (p.Gln174Ter)}. For that variant (chr15:90{,}749{,}788, hg38), \texttt{GeneBe\_\allowbreak{}classify\_\allowbreak{}variant} (API Tool) returns an independent ACMG/AMP verdict of \emph{Pathogenic} (score 12; criteria PVS1, PM2, PP5\_\allowbreak{}Moderate; rs1895622788; transcript NM\_000057.4). This is risk evidence, underscoring a genotoxicity concern for inhibition without validating \textit{BLM} as a target.
  \item \textbf{Test for a context-selective dependency.} \texttt{OpenTargets\_\allowbreak{}get\_\allowbreak{}target\_\allowbreak{}depmap\_\allowbreak{}essentiality} (API Tool) returns DepMap CRISPR gene-effect scores across 1{,}258 cancer cell lines. \textit{BLM} is not pan-essential (flagged non-essential, with a mean gene effect of $-0.18$ and only 7.5\% of lines, 94/1{,}258, dependent at $<-0.5$), yet its dependency concentrates in particular contexts: mature T- and NK-cell neoplasms ($-0.47$, $n{=}8$), cutaneous squamous-cell carcinoma ($-0.44$, $n{=}5$), and, at the tissue level, peripheral-nervous-system and neuroblastoma lines ($-0.33$, $n{=}48$). This is the turning point of the case: unlike the genetics, the screen reveals a quantitative, context-selective vulnerability that points to a possible therapeutic window in defined settings rather than to broad inhibition of a genome-stability factor.
  \item \textbf{Retrieve the structure and check its reliability.} \texttt{alphafold\_\allowbreak{}get\_\allowbreak{}summary} (API Tool) returns the model for the 1{,}417-residue protein (\texttt{AF-P54132-F1}) together with its confidence: a mean pLDDT of 60.5, indicating confidently folded catalytic regions interspersed with low-confidence interdomain linkers. Querying \texttt{PDBe\_\allowbreak{}get\_\allowbreak{}uniprot\_\allowbreak{}structure\_\allowbreak{}coverage} (API Tool) then surfaces high-resolution experimental structures of the helicase core, including PDB \texttt{7AUC} (X-ray, 1.53~\AA, residues $\sim$639--1290), \texttt{4CGZ} (a BLM--DNA complex, 3.2~\AA), and \texttt{4O3M} (a ternary complex, 2.3~\AA). Structure-based design should therefore be anchored on these experimental structures, with the AlphaFold model serving mainly to flag the unreliable linker regions.
  \item \textbf{Ground the therapeutic context in the literature.} \texttt{EuropePMC\_\allowbreak{}search\_\allowbreak{}articles} (API Tool) shows that the clinically validated synthetic-lethal target in the RecQ family is not \textit{BLM} but its paralog WRN, on which microsatellite-instable, mismatch-repair-deficient cancers depend, and that the WRN inhibitors HRO761 and VVD-133214 have entered clinical trials. The same search shows that ML216 itself has been used as a RecQ-helicase inhibitor with activity in MSI colorectal-cancer models. \textit{BLM} functions in the alternative lengthening of telomeres (ALT), consistent with the neuroblastoma-enriched dependency seen above, neuroblastoma being a canonical ALT-positive lineage. These results frame \textit{BLM} as a context-selective vulnerability hypothesis by analogy to the validated WRN/MSI precedent, with candidate contexts (mature T/NK neoplasms, cutaneous SCC, ALT-positive tumors) to test experimentally rather than a settled target.
  \item \textbf{Find and triage the chemical matter.} \texttt{OpenTargets\_\allowbreak{}get\_\allowbreak{}chemical\_\allowbreak{}probes\_\allowbreak{}by\_\allowbreak{}target\_\allowbreak{}ensemblID} (API Tool) identifies ML216 but flags it as low quality (\texttt{isHighQuality} = false; ProbeMiner score 20; experimental origin), so the platform itself marks it a probe rather than a lead. \texttt{PubChem\_\allowbreak{}get\_\allowbreak{}CID\_\allowbreak{}by\_\allowbreak{}compound\_\allowbreak{}name} and \texttt{PubChem\_\allowbreak{}get\_\allowbreak{}compound\_\allowbreak{}properties\_\allowbreak{}by\_\allowbreak{}CID} (API Tools) resolve its structure (CID 49852229, a urea--thiadiazole of MW 383~Da), and \texttt{ADMETAI\_\allowbreak{}predict\_\allowbreak{}physicochemical\_\allowbreak{}properties} and \texttt{ADMETAI\_\allowbreak{}predict\_\allowbreak{}toxicity} (ML Model Tools) find it drug-like (QED 0.66, no Lipinski violations) but flag a high predicted risk of drug-induced liver injury (DILI 0.99) alongside moderate hERG liability (0.56). Profiling five close analogs (\texttt{ChEMBL\_\allowbreak{}search\_\allowbreak{}similar\_\allowbreak{}molecules}, 72--78\% similarity), predicted hERG (0.45--0.67) and mutagenicity (AMES 0.12--0.31) vary across the series, whereas DILI remains near 0.99 (0.987--0.992) for every analog. This pattern is consistent with a scaffold-associated DILI alert; it is a model-based hypothesis that would need experimental hepatotoxicity data to confirm, and if real would call for a scaffold change rather than peripheral substitution.
\end{enumerate}

\begin{center}
\small
\begin{tabular}{@{}lll@{}}
\toprule
\textbf{ADMET-AI prediction (ML216)} & \textbf{Value} & \textbf{Interpretation} \\
\midrule
Molecular weight & 383 Da & within drug-like range \\
QED (drug-likeness) & 0.66 & favorable \\
Lipinski violations & 0 of 4 & passes rule of five \\
DILI (liver injury) & 0.99 & \textbf{high predicted risk---flag} \\
hERG (cardiac) & 0.56 & moderate predicted risk \\
AMES (mutagenicity) & 0.16 & low predicted risk \\
DILI across 5 close analogs & 0.987--0.992 & persistent; scaffold-associated (model hypothesis) \\
\bottomrule
\end{tabular}
\end{center}

\noindent \textbf{Outcome.} \textit{BLM} is a genetics-validated cancer-predisposition gene whose loss supports a genome-maintenance role, so inhibition is not warranted by the genetics, which instead raise a genotoxicity concern. The evidence is better read as a context-selective vulnerability hypothesis: DepMap shows a modest dependency in defined settings (mature T/NK-cell neoplasms, cutaneous squamous-cell carcinoma, and neuroblastoma or ALT-positive lines), and the literature supplies a validated RecQ-family precedent in the paralog WRN for MSI/MMR-deficient cancers, together prioritizing specific contexts and biomarkers for experimental testing rather than establishing a target. ML216 is drug-like but platform-flagged as low quality, and its predicted liver-injury liability persists across close analogs, making it an early chemical probe rather than a developable lead. The value is a calibrated synthesis that keeps the safety, opportunity, and chemical-liability signals distinct and reconciles them into one testable hypothesis, each claim traceable to its tool.

\subsection{Case 2: Assessing \textit{OXTR} druggability and the chemical strategy for a CNS indication}\label{case:oxtr}
\emph{Research question: ``Is \textit{OXTR} (the oxytocin receptor) a druggable target, and does its pharmacology offer a credible chemical starting point for its emerging autism association?''} The AI scientist assembles a target dossier, finding that the straightforward ``activate the receptor'' reading needs qualification at several points.

\begin{enumerate}
  \item \textbf{Resolve, annotate, and obtain structure.} \texttt{OpenTargets\_\allowbreak{}multi\_\allowbreak{}entity\_\allowbreak{}search\_\allowbreak{}by\_\allowbreak{}query\_\allowbreak{}string} maps \textit{OXTR} to \texttt{ENSG00000180914}, and \texttt{OpenTargets\_\allowbreak{}get\_\allowbreak{}target\_\allowbreak{}gene\_\allowbreak{}ontology\_\allowbreak{}by\_\allowbreak{}ensemblID} identifies a plasma-membrane G-protein-coupled receptor that signals through a phospholipase-C/calcium pathway (GO:0004990, GO:0007200). \texttt{alphafold\_\allowbreak{}get\_\allowbreak{}summary} returns a confidently modeled transmembrane core (\texttt{AF-P30559-F1}, 389 residues, mean pLDDT 78.6), but \texttt{PDBe\_\allowbreak{}get\_\allowbreak{}uniprot\_\allowbreak{}structure\_\allowbreak{}coverage} (API Tools) surfaces experimental structures in both functional states: an agonist-bound active complex (PDB \texttt{7RYC}, cryo-EM 2.9~\AA, \textit{OXTR} bound to oxytocin with a heterotrimeric G$_{\mathrm q}$ protein, and \texttt{7QVM} at 3.25~\AA) and an inactive receptor (\texttt{6TPK}, X-ray 3.2~\AA). Because GPCR design depends on ligand-bound, state-specific conformations, these experimental structures rather than the predicted model are the appropriate basis for design, and the active, agonist-bound state matches the pharmacology the indication requires.
  \item \textbf{Map the disease landscape.} \texttt{OpenTargets\_\allowbreak{}get\_\allowbreak{}diseases\_\allowbreak{}phenotypes\_\allowbreak{}by\_\allowbreak{}target\_\allowbreak{}ensembl} (API Tool) returns 393 associations, spanning reproductive and uterine outcomes with approved-drug precedent as well as a distinct association with autism spectrum disorder. The latter is an emerging, hypothesis-generating direction rather than a clinically validated one.
  \item \textbf{Confirm tractability and enumerate drugs.} \texttt{OpenTargets\_\allowbreak{}get\_\allowbreak{}target\_\allowbreak{}tractability\_\allowbreak{}by\_\allowbreak{}ensemblID} classifies \textit{OXTR} as a druggable-family GPCR with a ligand-bound structure and high-quality ligands, but its approved agents are peptides rather than small molecules, and non-peptide chemistry reaches only advanced-clinical stage. \texttt{OpenTargets\_\allowbreak{}get\_\allowbreak{}associated\_\allowbreak{}drugs\_\allowbreak{}by\_\allowbreak{}target\_\allowbreak{}ensemblID} (API Tools) returns nine agents, among them oxytocin, atosiban, and carbetocin (approved), retosiban and nolasiban (phase 3), and barusiban and epelsiban (phase 2). This establishes ligand precedent for reproductive and uterine indications, which does not de-risk an autism indication.
  \item \textbf{Resolve mechanism.} \texttt{OpenTargets\_\allowbreak{}get\_\allowbreak{}drug\_\allowbreak{}mechanisms\_\allowbreak{}of\_\allowbreak{}action\_\allowbreak{}by\_\allowbreak{}chemblId} (API Tool) shows the receptor is drugged in both directions: oxytocin is an agonist that induces labor, and atosiban an antagonist used as a tocolytic.
  \item \textbf{Confirm chemical matter with experimental data.} Resolving the ChEMBL target (\texttt{ChEMBL\_\allowbreak{}search\_\allowbreak{}targets}~$\rightarrow$~CHEMBL2049) and retrieving its activities (\texttt{ChEMBL\_\allowbreak{}get\_\allowbreak{}target\_\allowbreak{}activities}, API Tools) returns measured binding affinities reaching single-digit nanomolar potency ($K_i$ as low as 3.2~nM), confirming that the receptor has well-characterized, high-affinity ligands.
  \item \textbf{Reason about pharmacology and CNS druggability.} The pro-social hypothesis calls for receptor activation, but the clinical evidence is mixed. \texttt{EuropePMC\_\allowbreak{}search\_\allowbreak{}articles} shows that no oxytocin product is approved for the core social-communication symptoms of autism, that intranasal oxytocin remains investigational in recent reviews, and that trials are confounded by large placebo responses, including a randomized lead-in in which nearly half of participants met placebo-responder criteria. The medicinal-chemistry objective is therefore not simply ``agonist good, antagonist bad'' but a brain-penetrant agonist, partial agonist, or positive allosteric modulator with selectivity over the vasopressin receptors. To test whether a brain-penetrant non-peptide chemotype is even attainable, the workflow profiles nolasiban (\texttt{PubChem\_\allowbreak{}get\_\allowbreak{}CID\_\allowbreak{}by\_\allowbreak{}compound\_\allowbreak{}name}, CID 52947354) with \texttt{ADMETAI\_\allowbreak{}predict\_\allowbreak{}physicochemical\_\allowbreak{}properties} and \texttt{ADMETAI\_\allowbreak{}predict\_\allowbreak{}BBB\_\allowbreak{}penetrance} (ML Model Tools), finding it drug-like (QED 0.87) with high predicted blood-brain-barrier penetrance (0.93). That figure indicates plausibility, not proven central exposure, and in any case nolasiban is an antagonist: brain-penetrant non-peptide \textit{OXTR} chemistry is feasible, but the available CNS-capable chemotype blocks the receptor rather than activating it.
\end{enumerate}
\noindent \textbf{Outcome.} Rather than a naive ``repurpose an approved drug'' suggestion, \toun delivers a mechanistically careful conclusion. \textit{OXTR} is a tractable, well-liganded GPCR with experimental structures in both active and inactive states, clinically validated for reproductive and uterine indications, whereas its autism association remains an emerging hypothesis with inconsistent clinical support. Central engagement is chemically plausible, but the pro-social hypothesis requires agonism while the only brain-penetrant chemotype on hand is an antagonist. Integrating mechanism, experimental affinity and structure, and machine-learning prediction, the platform defines a precise objective---a brain-penetrant \textit{OXTR} agonist or positive allosteric modulator, selective over the vasopressin receptors and with confirmed central exposure---and so avoids the pharmacologically wrong repurposing call a drug-name lookup would invite.

\subsection{Case 3: Clinical-trial safety biostatistics for BCG vaccination and adverse-event severity}\label{case:bcg}
\emph{Research question: ``In the public BCG-CORONA healthcare-worker trial, is BCG vaccination associated with higher adverse-event severity after adjusting for patient-interaction frequency?''} Unlike Cases 1 and 2, this question calls for statistical computation on raw tabular data. The AI scientist uses \texttt{clinical\_\allowbreak{}trial\_\allowbreak{}ae\_\allowbreak{}severity\_\allowbreak{}test} (Software Tool), which merges the trial's demographics (DM) and adverse-events (AE) tables and models severity by treatment arm. Following the tool's convention, each subject's outcome is the maximum severity grade (AESEV) across all of their adverse events, and the cohort is those subjects with at least one adverse-event record, obtained by an inner join of AE onto DM. Data are from the publicly available BCG-CORONA trial (\url{https://zenodo.org/records/12737228}).

\begin{enumerate}
  \item \textbf{Construct the analysis cohort} (\texttt{prepare} mode): the inner join yields 791 subjects with evaluable AE records, each reduced to a single maximum-severity value (distribution \{grade 1:~328, 2:~402, 3:~43, 4:~18\}).
  \item \textbf{Test for an unadjusted association} (\texttt{chi-square} mode): the treatment-by-severity table shows a significant association ($\chi^{2}=10.12$, $\mathrm{dof}=3$, $p=0.018$).
  \item \textbf{Fit an adjusted model} (\texttt{ordinal} mode) with patient-interaction covariates: a multivariable ordinal logistic (proportional-odds) regression gives an odds ratio of $1.53$ (95\% CI $1.16$--$2.01$, $p=0.0024$) for higher severity with BCG vaccination, and the covariates are not significant. Because this is a proportional-odds model, the single odds ratio assumes a common effect across severity thresholds, an assumption that should be checked before the estimate is over-interpreted.
\end{enumerate}

\begin{center}
\small
\begin{tabular}{@{}lll@{}}
\toprule
\textbf{Step / test} & \textbf{Tool mode} & \textbf{Result} \\
\midrule
Cohort construction & \texttt{prepare} & 791 AE-evaluable subjects; AESEV \{1:328, 2:402, 3:43, 4:18\} \\
Unadjusted association & \texttt{chi-square} & $\chi^{2}=10.12$, $\mathrm{dof}=3$, $p=0.018$ \\
Adjusted model & \texttt{ordinal} & OR $=1.53$ (95\% CI $1.16$--$2.01$), $p=0.0024$ \\
\bottomrule
\end{tabular}
\end{center}

\noindent \textbf{Outcome.} The unadjusted test and the adjusted model concur that, in this cohort, BCG assignment is associated with higher maximum adverse-event severity (adjusted OR 1.53); the two are not independent, since they share the outcome and overlapping data. Because the cohort is restricted to participants with adverse-event records and the endpoint is a derived per-subject maximum, the finding is best read as a reproducible safety signal from a secondary analysis rather than a definitive effect. As a demonstration, \toun performs the analysis---cohort construction, endpoint derivation, and modeling---directly on raw trial tables, making the cohort and endpoint definitions explicit.

\subsection{Strengths and limitations illustrated by the case studies}
\emph{Strengths.}
\begin{itemize}
\item \textbf{From a natural-language question, end to end.} Each case begins with a plain-language question that the AI scientist answers by autonomously selecting and chaining tools, reporting a concrete result at each step without user-written code.
\item \textbf{Composition across modalities.} One workflow spans human genetics, functional-genomics screens (DepMap), protein structure (AlphaFold and experimental PDB structures), cheminformatics and experimental bioactivity (ChEMBL), machine-learning prediction (ADMET-AI), and statistics, assembling in a single run the evidence that would otherwise demand many separate resources and custom code.
\item \textbf{Reconciling conflicting evidence.} The value is adjudication, not retrieval: in the \textit{BLM} case the workflow holds a safety signal, an opportunity signal, and a predicted chemical liability apart, then reconciles them into a testable, biomarker-defined hypothesis rather than treating a predisposition gene as an automatic target.
\item \textbf{Calibrated conclusions.} It avoids overclaiming, returning \textit{BLM} as a hypothesis rather than a target, superseding a low-confidence predicted structure with experimental ones, treating a predicted BBB score as plausibility rather than exposure, and reporting a persistent predicted-DILI pattern as a model hypothesis to confirm.
\item \textbf{Reproducibility and provenance.} Every step is a typed call through one interface, so each result carries the name and arguments of the tool that produced it; with versioned releases and optional audit logging, a workflow can be inspected and re-run.
\item \textbf{Cross-checking that catches errors.} Combining evidence types exposes what a single lookup misses: in the \textit{OXTR} case, a brain-penetrant hit (nolasiban) looks like a repurposing candidate until its mechanism reveals an antagonist where the hypothesis needs activation, a mismatch the workflow flags before it becomes a recommendation.
\end{itemize}

\emph{Limitations.} \toun orchestrates rather than certifies: it composes existing tools rather than re-deriving them, so any conclusion is bounded by the underlying databases and models. Machine-learning outputs such as an ADMET-AI score, and association evidence, are estimates to weigh by provenance and validation domain and confirm experimentally, not measurements; a dependency score prioritizes tumor contexts but does not establish pharmacologic tractability, tumor-selective toxicity, or an adequate therapeutic index; and quantitative results still need domain judgment, including whether a derived endpoint captures the intended construct (a per-subject maximum over all adverse events), which cohort an inner join defines, and an odds ratio's direction relative to its reference level (reported as $0.65$ or its reciprocal $1.53$). \toun mitigates these with a human-expert curation gate, recorded provenance, and human-in-the-loop checkpoints, and it makes the evidence, computations, and predictions explicit so these judgments are transparent. It does not by itself certify scientific correctness, and the scientist remains in the loop.

\fussy

\clearpage

\begin{figure}[ht]
\begin{promptsllm}[Tool Description Schema in \toun]
    \item \textbf{Name}: The unique identifier for the tool.
    \item \textbf{Description}: A clear and concise summary of the tool's purpose and functionality.
    \item \textbf{Parameters}: A list of arguments that the tool accepts. Each argument has the following properties:
    \begin{adjustwidth}{1em}{1em}
        \begin{itemize}
            \item \textbf{Argument Name}: The name of the parameter.
            \item \textbf{Argument Type}: The expected data type for the parameter's value (e.g., string, integer, boolean).
            \item \textbf{Argument Description}: A detailed explanation of what the parameter represents and its purpose.
            \item \textbf{Required}: A boolean value indicating whether the parameter is mandatory for the tool to execute.
        \end{itemize}
    \end{adjustwidth}
    \item \textbf{Return Schema}: A description of the structure and data types of the output returned by the tool upon successful execution.
\end{promptsllm}
\caption{The tool description schema in \toun is consistent across all tools, regardless of their diverse backends.}
\label{fig:tool_spec}
\end{figure}

\begin{figure}[ht]
\centering
\includegraphics[width=1.0\textwidth]{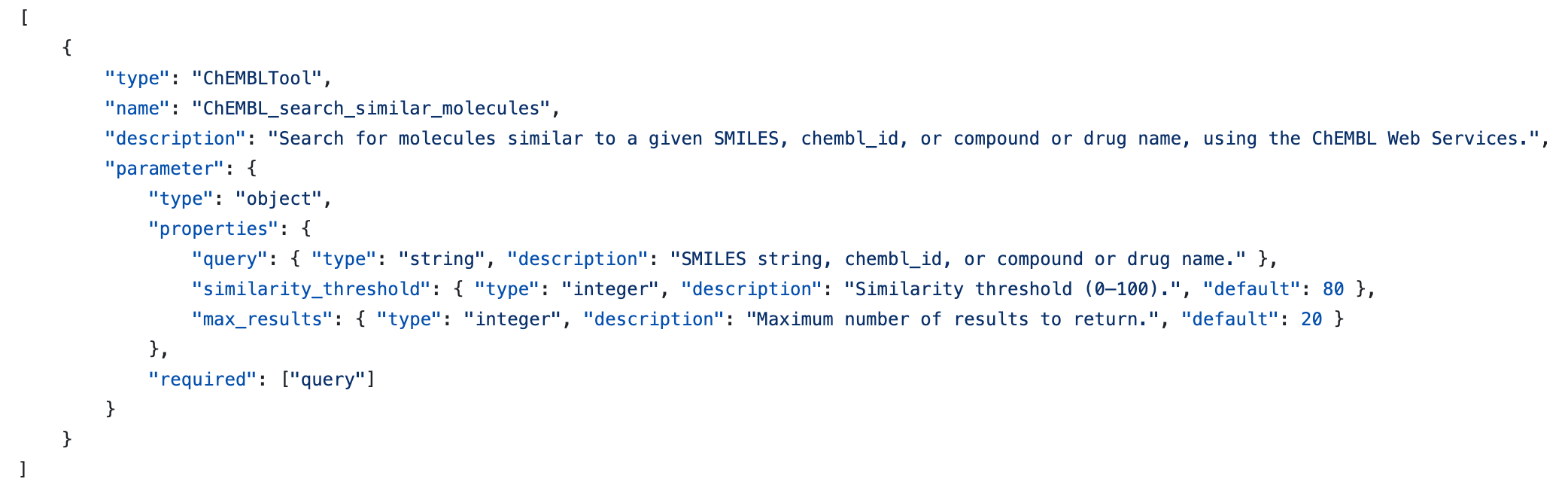}
\caption{One example of a tool description in \toun.}
\label{fig:tool_example}
\end{figure}

\begin{figure}[htbp]
\begin{promptsllm}[Interaction schema in \toun]
    \item \textbf{Name}: The name of the tool or operation to be called.
    \item \textbf{Parameters}: A list of arguments that the tool accepts. Each argument has the following properties:
    \begin{adjustwidth}{1em}{1em}
        \begin{itemize}
            \item \textbf{Argument Name}: The name of the parameter.
            \item \textbf{Argument Value}: The value for the parameter provided by the client.
        \end{itemize}
    \end{adjustwidth}
\end{promptsllm}
\caption{The universal interaction schema for all tools and operations within \toun.}
\label{fig:tool_interact}
\end{figure}

\begin{figure}[t]
\centering
\includegraphics[width=1.0\textwidth]{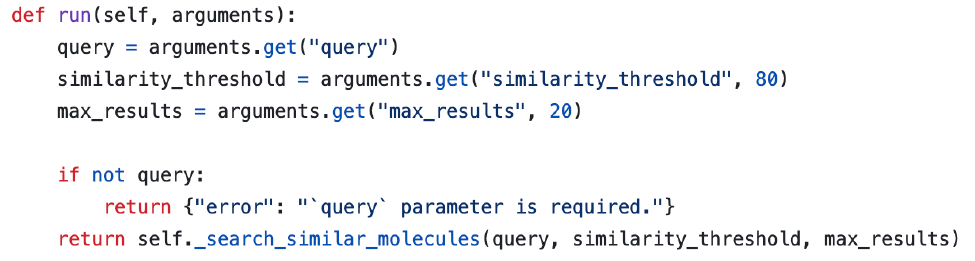}
\caption{Code example of a tool operation implementation used by the Tool Caller during execution.}
\label{fig:run_tool_example}
\end{figure}

\begin{figure}[t]
\centering
\includegraphics[width=0.9\textwidth]{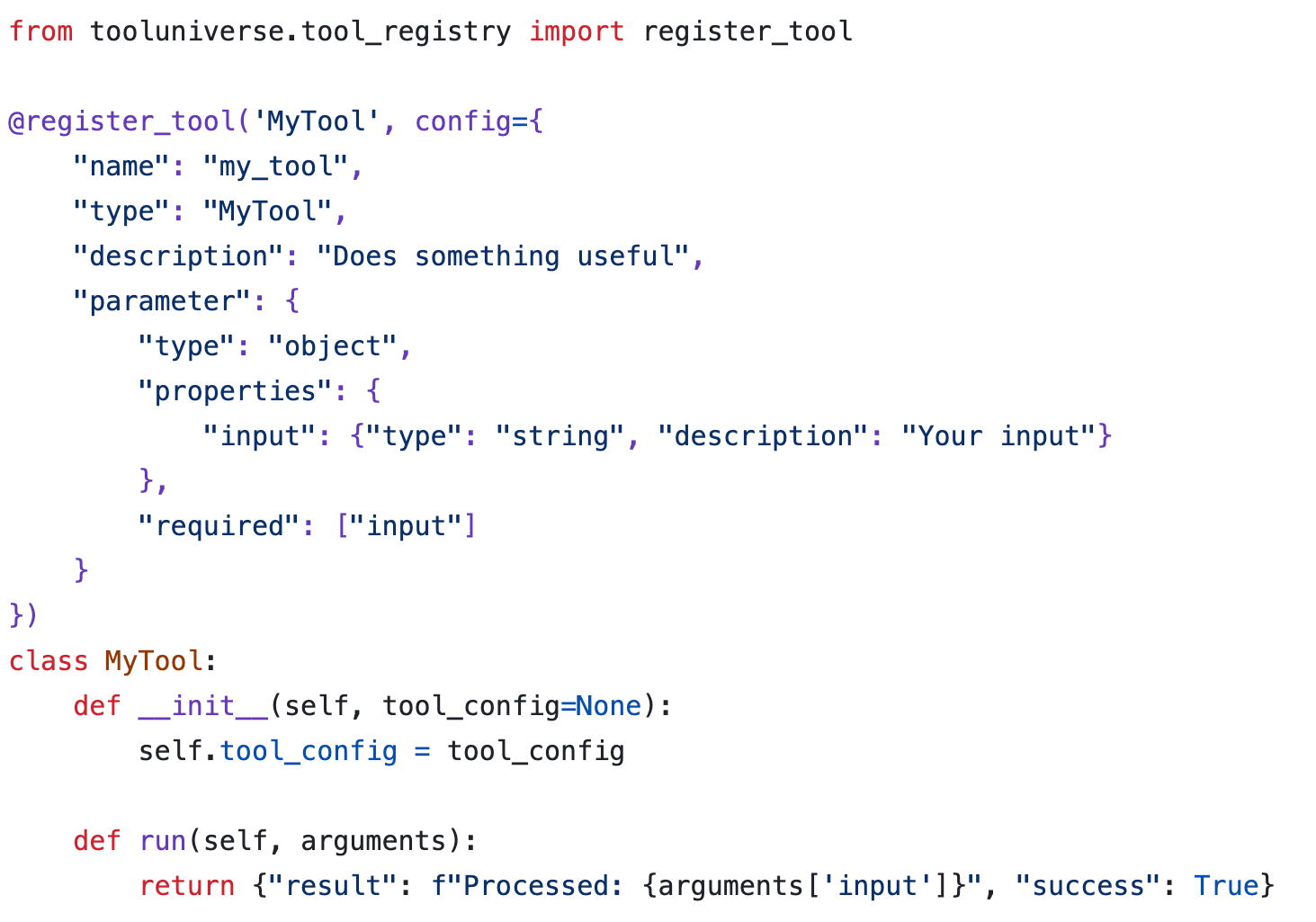}
\caption{Example code demonstrating how to register a local tool with Tool Manager and add it to \toun.}
\label{fig:tool_local_reg}
\end{figure}

\begin{figure}[t]
\centering
\includegraphics[width=0.9\textwidth]{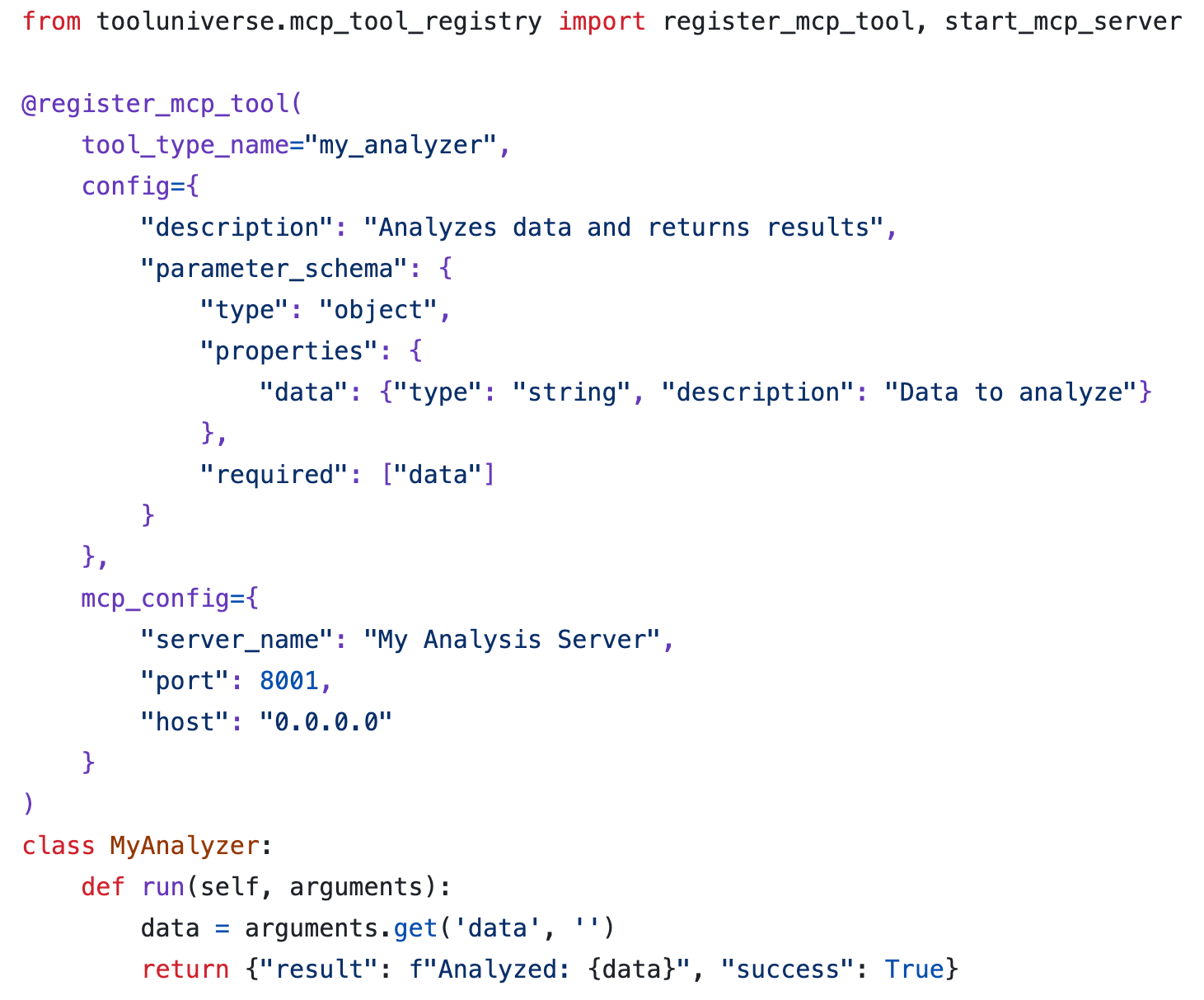}
\caption{Example code demonstrating how to register a remote tool with Tool Manager and add it to \toun.}
\label{fig:tool_remote_reg}
\end{figure}

\begin{figure}[ht]
\centering
\includegraphics[width=1.0\textwidth]{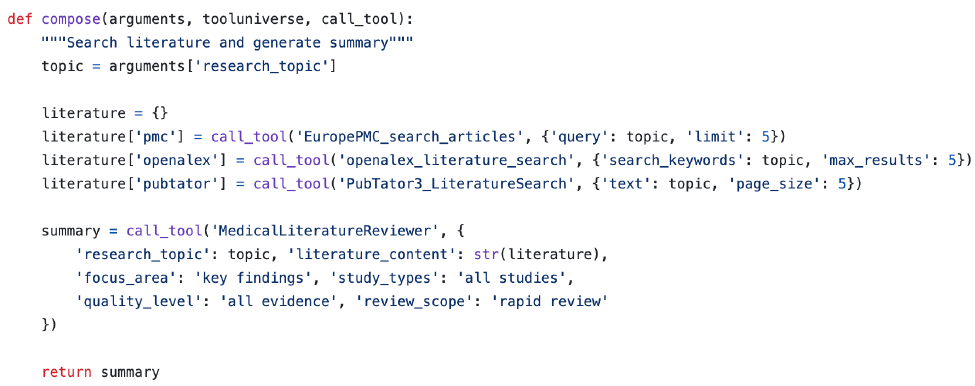}
\caption{This example demonstrates a composable tool, built with the Tool Composer, that runs multiple literature search tools concurrently, followed by a summary agent that synthesizes the results. The Tool Composer enables the combination of multiple tools from \toun in diverse ways, such as in parallel, sequentially, or in loops, enabling multi-tool collaboration.}
\label{fig:tool_compose}
\end{figure}

\clearpage

\setlength{\tabcolsep}{2mm}
\begin{longtable}[t]{>{\raggedright\arraybackslash}p{0.18\columnwidth}|>{\raggedright\arraybackslash}p{0.75\columnwidth}}
\toprule
\textbf{Tool category} & \textbf{Description} \\
 \midrule
ML models & Tools that apply machine learning algorithms to tasks like prediction, classification, or generation. Examples include foundation models, language models, and reasoning models.\\
\midrule
Agents & Tools that operate autonomously to perceive environments, make decisions, and take actions toward goals. Examples include research planning agents, hypothesis generation agents, and data analysis agents.\\
\midrule
Scientific packages & Software packages engineered to facilitate diverse scientific tasks, experiments, and data analysis workflows. Examples include computation packages, biomedical libraries, and scientific simulators. \\
\midrule
Automation & Tools involving physical or simulated machines capable of sensing, reasoning, and acting in the world. Examples include lab automation tools and instrument control policies.\\
\midrule
Workflows & Tools that enable complex, multi-step scientific workflows. Examples include orchestration schedulers and RAG pipelines. \\
\midrule
Visualization & Tools that facilitate the display and communication of scientific data and results. Examples include dashboards and plot and chart generators. \\
\midrule
Feedback and safety & Tools that incorporate evaluations or input from human experts to ensure safety and goal alignment. Examples include privacy guardrails, security checklists, and human-in-the-loop tools. \\
\midrule
Local and remote datasets & Tools that store, manage, and query structured or semi-structured data, including relational, tabular, and hierarchical datasets. Examples include knowledge bases and typed API clients. \\
\midrule
Experimentation & Tools that support the design and management of experiments. Examples include protocol generators and LIMS integration tools. \\
\midrule
Search and embeddings & Tools that store and retrieve vectorized representations of data for use in machine learning tasks. Examples include vector search tools and embedding generators. \\
\bottomrule
\caption{Types of tools in \toun that \toun-powered AI scientists can use.}\label{table:tool_type}
\end{longtable}

\clearpage

\begingroup
\setlength{\tabcolsep}{2mm}
\centering
\footnotesize\let\small\footnotesize
\begin{tabular}{>{\raggedright\arraybackslash}p{0.17\columnwidth}|p{0.09\columnwidth}|>{\raggedright\arraybackslash}p{0.66\columnwidth}}
\toprule
\textbf{Tool category} & \textbf{Number} & \textbf{Examples of tools in \toun}\\
 \midrule
ML models & 89 &  \makecell[l]{\small\texttt{ADMETAI\_predict\_toxicity}, \\ \small\texttt{SpliceAI\_predict\_splice},  \\ \small\texttt{IEDB\_predict\_mhci\_binding}}  \\
\midrule
Agents & 45 & \makecell[l]{\small\texttt{HypothesisGenerator},   \\ \small\texttt{CodeQualityAnalyzer},  \\ \small\texttt{MedicalLiteratureReviewer}} \\
\midrule
Scientific packages & 179 & \makecell[l]{\small\texttt{get\_biopython\_info}, \\ \small\texttt{get\_rdkit\_info}, \\   \small\texttt{get\_scanpy\_info}}\\
\midrule
Automation & 4 &  \makecell[l]
{
\small\texttt{mcp\_auto\_loader\_boltz}, \\
\small\texttt{mcp\_auto\_loader\_txagent}, \\
\small\texttt{mcp\_auto\_loader\_esm} } \\
\midrule
Workflows & 16 &  \makecell[l]
{
\small\texttt{BiomarkerDiscoveryWorkflow} \\
\small\texttt{ComprehensiveDrugDiscoveryPipeline} \\
\small\texttt{ToolDiscover}
} \\
\midrule
Visualization & 3  &  \makecell[l]
{
\small\texttt{visualize\_protein\_structure\_3d} \\
\small\texttt{visualize\_molecule\_2d} \\
\small\texttt{visualize\_molecule\_3d}
} \\
\midrule
Feedback and safety & 7 &  \makecell[l]{\small\texttt{consult\_human\_expert}, \\ \small\texttt{get\_expert\_response}, \\ \small\texttt{get\_expert\_status}} \\
\midrule
Local and remote datasets & 2,421 &  \makecell[l]
{\small\texttt{ChEMBL\_get\_molecule}, \\
\small\texttt{HPA\_get\_comprehensive\_gene\_details\_by\_ensembl\_id}, \\
\small\texttt{FDA\_get\_active\_ingredient\_info\_by\_drug\_name},  \\
\small\texttt{OpenTargets\_get\_associated\_targets\_by\_disease\_efoId}} \\
\midrule
Search and embeddings & 9 & \makecell[l]{\small\texttt{Tool\_Finder}, \\ \small\texttt{Tool\_RAG}, \\ \small\texttt{embedding\_database\_search}}\\
\midrule
Experimentation & 4 & \makecell[l]{\small\texttt{ExperimentalDesignScorer}, \\ \small\texttt{ProtocolOptimizer}, \\ \small\texttt{extract\_clinical\_trial\_outcomes}}\\
\midrule
\midrule
\textbf{Unique tools} & \textbf{2,777} & \makecell[l]{\small\textit{Total number of unique tools across all categories}}\\
\bottomrule
\end{tabular}
\par
\normalsize
\captionof{table}{Number of tools and example tools for each category in \mbox{\toun}, from a direct enumeration of the open-source repository (\texttt{main} branch) on 7 July 2026. Each tool is assigned to a single primary category from its implementation type, name, and source, so the category counts sum to the total of 2,777 unique tools. \mbox{\toun} is continually expanded with new tools.}
\label{table:tool_example}
\par
\endgroup

To characterize coverage of the tool library by data modality and application area, Supplementary Table~\ref{table:tool_modality} reports the number of tools in each area. Because a tool may serve several modalities, it is counted in every area to which it applies.

\begin{longtable}[t]
{>{\raggedright\arraybackslash}p{0.36\columnwidth}|p{0.07\columnwidth}|>{\raggedright\arraybackslash}p{0.49\columnwidth}}
\toprule
\textbf{Data-modality / application area} & \textbf{Tools} & \textbf{Representative resources}\\
\midrule
Drugs \& pharmacology & 474 & \small FDA drug labels, PharmGKB, DailyMed\\
\midrule
Clinical, disease \& phenotypes & 289 & \small ClinicalTrials.gov, Open Targets, HPO / Orphanet\\
\midrule
Genomes, sequences \& browsers & 222 & \small Ensembl, UCSC Genome Browser, ENA\\
\midrule
Proteins, sequence, function \& proteomics & 193 & \small UniProt, InterPro, PRIDE\\
\midrule
Tooling, methods \& infrastructure & 193 & \small Agentic tools, Tool Finder, software packages\\
\midrule
Macromolecular structure \& prediction & 189 & \small RCSB PDB, AlphaFold, PDBe\\
\midrule
Genetic variation, GWAS \& population genetics & 184 & \small gnomAD, GWAS Catalog, ClinVar\\
\midrule
Pathways \& molecular networks & 169 & \small Reactome, KEGG, STRING\\
\midrule
Chemistry \& small molecules & 156 & \small ChEMBL, PubChem, ADMET-AI\\
\midrule
Literature \& scholarly data & 143 & \small PubMed, Europe PMC, OpenAlex\\
\midrule
Transcriptomics \& gene expression & 137 & \small GTEx, GEO, Human Protein Atlas\\
\midrule
Earth, space, ecology \& socioeconomic & 120 & \small NASA, Open-Meteo, GBIF\\
\midrule
Annotation, ontologies \& enrichment & 111 & \small Gene Ontology, Enrichr, OLS\\
\midrule
Cancer genomics \& oncology & 106 & \small cBioPortal, GDC, OncoKB\\
\midrule
Model organisms \& comparative genomics & 87 & \small SGD, ZFIN, Alliance of Genome Resources\\
\midrule
Metabolomics \& lipidomics & 78 & \small MetaboLights, GNPS, LIPID MAPS\\
\midrule
Epigenomics \& gene regulation & 78 & \small ENCODE, JASPAR, ChIP-Atlas\\
\midrule
Imaging \& neuroscience & 77 & \small CryoET Portal, OpenNeuro, Allen Brain\\
\midrule
Microbiome \& infectious disease & 54 & \small MGnify, GTDB, BV-BRC\\
\midrule
Immunology & 43 & \small IEDB, VDJdb, IMGT\\
\midrule
\bottomrule
\caption{Coverage of the \toun tool library by data modality and application area (open-source repository \texttt{main} branch, 7 July 2026). Categories are not mutually exclusive: a tool serving several modalities is counted in each, so the column sums to more than the number of unique tools (2{,}777 unique; 3{,}103 category memberships). Assignments were generated programmatically from each tool's resource and description. Supplementary Table~\ref{table:skill_categories} lists 130 research skills, grouped by application areas and data modalities.} \label{table:tool_modality}
\end{longtable}

\begingroup
\catcode`\-=\active
\def-{\char`\-\penalty0\hskip0pt\relax}
\footnotesize
\begin{longtable}[t]{>{\raggedright\arraybackslash}p{0.34\columnwidth}|>{\raggedright\arraybackslash}p{0.60\columnwidth}}
\toprule
\textbf{Skill} & \textbf{Description}\\ \midrule \endhead
\multicolumn{2}{l}{\cellcolor{gray!15}\textbf{Genomics, sequences and variants}}\\ \midrule
\texttt{\small tooluniverse-acmg-variant-classification} & \small Systematic ACMG/AMP germline variant classification with all 28 criteria (PVS1, PS1-4, PM1-6, PP1-5, BA1, BS1-4, BP1-7) for...\\
\texttt{\small tooluniverse-variant-analysis} & \small VCF and variant analysis: parsing, annotation, classification (synonymous, missense, frameshift, stop\_gained), VAF filtering,...\\
\texttt{\small tooluniverse-variant-interpretation} & \small Clinical variant interpretation from raw variant calls to ACMG-classified recommendations with structural impact analysis\\
\texttt{\small tooluniverse-variant-functional-annotation} & \small Functional annotation of protein variants: ProtVar structural/functional context, ClinVar clinical classifications, gnomAD...\\
\texttt{\small tooluniverse-variant-to-mechanism} & \small End-to-end variant-to-mechanism analysis: trace a variant (rsID/coordinates) through regulatory context, target gene(s), molecular...\\
\texttt{\small tooluniverse-structural-variant-analysis} & \small Structural variant (SV) clinical interpretation: deletions, duplications, inversions, translocations, complex rearrangements\\
\texttt{\small tooluniverse-regulatory-variant-analysis} & \small Non-coding/regulatory variant interpretation: GWAS association lookup, eQTL evidence (GTEx), chromatin state (ENCODE), regulatory...\\
\texttt{\small tooluniverse-cancer-variant-interpretation} & \small Clinical interpretation of somatic cancer mutations for precision oncology\\
\texttt{\small tooluniverse-sequence-analysis} & \small Biological sequence analysis: gene/protein sequence retrieval (NCBI, Ensembl, UniProt), nucleotide/protein search, ortholog...\\
\texttt{\small tooluniverse-sequence-retrieval} & \small Retrieve DNA/RNA/protein sequences from NCBI and ENA with disambiguation\\
\texttt{\small tooluniverse-fastq-qc} & \small FASTQ quality control and adapter/quality-trimming decisions with local NGS tools: run FastQC on raw reads, summarize a project...\\
\texttt{\small tooluniverse-primer-design} & \small PCR / qPCR primer and oligo design: design forward/reverse primers for a target region (SantaLucia nearest-neighbor...\\
\texttt{\small tooluniverse-molecular-cloning} & \small Molecular cloning assembly design: Gibson Assembly (overlap design for seamless multi-fragment joining) and Golden Gate Assembly...\\
\texttt{\small tooluniverse-comparative-genomics} & \small Cross-species gene comparison and ortholog analysis\\
\texttt{\small tooluniverse-model-organism-genetics} & \small Cross-species genetic analysis using model organism databases (MGI mouse, ZFIN zebrafish, FlyBase fruit fly, WormBase worm, SGD...\\
\texttt{\small tooluniverse-plant-genomics} & \small Plant genomics and biology research: PlantReactome pathways, Ensembl Plants gene structure, POWO species taxonomy, UniProt...\\
\texttt{\small tooluniverse-rare-disease-genomics} & \small Rare disease genomics: disease identification (Orphanet), causative gene discovery, gene-disease validity (GenCC), variant...\\
\texttt{\small tooluniverse-cancer-genomics-tcga} & \small TCGA/GDC cancer genomics analysis: cohort construction, clinical metadata retrieval, somatic mutation frequencies, survival...\\
\texttt{\small tooluniverse-microbial-genome-characterization} & \small Genome-ASSEMBLY discovery, QC, and replicon mapping for any organism (bacteria, archaea, fungi, and beyond) using NCBI Datasets\\
\midrule
\multicolumn{2}{l}{\cellcolor{gray!15}\textbf{GWAS, population and statistical genetics}}\\ \midrule
\texttt{\small tooluniverse-gwas-drug-discovery} & \small Transform GWAS signals into drug targets and repurposing opportunities\\
\texttt{\small tooluniverse-gwas-finemapping} & \small Statistical fine-mapping of GWAS loci using credible sets (SuSiE, FINEMAP) and locus-to-gene scoring (Open Targets L2G)\\
\texttt{\small tooluniverse-gwas-snp-interpretation} & \small Interpret a single GWAS SNP across multiple databases: GWAS Catalog hits, LD/haplotype context, eQTL evidence, regulatory...\\
\texttt{\small tooluniverse-gwas-study-explorer} & \small Compare GWAS studies, perform meta-analyses across cohorts, and assess signal replication\\
\texttt{\small tooluniverse-gwas-trait-to-gene} & \small Discover causal genes for diseases/traits from GWAS data using Open Targets L2G (locus-to-gene) scoring: integrates eQTL,...\\
\texttt{\small tooluniverse-phewas} & \small Cross-ancestry / cross-biobank phenome-wide association (PheWAS) and replication\\
\texttt{\small tooluniverse-polygenic-risk-score} & \small Build and interpret polygenic risk scores (PRS) for complex diseases using GWAS summary statistics\\
\texttt{\small tooluniverse-population-genetics} & \small Population genetics analysis: allele frequencies (gnomAD, 1000 Genomes), Hardy-Weinberg equilibrium testing, Fst between...\\
\texttt{\small tooluniverse-population-genetics-1000genomes} & \small Population genetics using the 1000 Genomes Project (IGSR): superpopulation/population search, sample metadata, variant frequencies...\\
\texttt{\small tooluniverse-mendelian-randomization} & \small Mendelian randomization (MR) causal inference: does an exposure, risk factor, or biomarker CAUSALLY affect a disease/outcome,...\\
\texttt{\small tooluniverse-gene-disease-association} & \small Gene-disease association analysis across DisGeNET, OpenTargets, Monarch, OMIM, GenCC, Orphanet\\
\texttt{\small tooluniverse-pathway-disease-genetics} & \small Connect GWAS variants to biological pathways and druggable targets\\
\midrule
\multicolumn{2}{l}{\cellcolor{gray!15}\textbf{Transcriptomics and functional genomics}}\\ \midrule
\texttt{\small tooluniverse-rnaseq-deseq2} & \small RNA-seq differential expression analysis with DESeq2, edgeR, and limma-voom: DEG lists, fold changes, dispersion estimation,...\\
\texttt{\small tooluniverse-single-cell} & \small Single-cell RNA-seq analysis with scanpy/anndata: h5ad data loading, scRNA-seq quality control and QC gating (n\_genes\_by\_counts,...\\
\texttt{\small tooluniverse-spatial-transcriptomics} & \small Spatial transcriptomics analysis: Visium, MERFISH, seqFISH, Slide-seq\\
\texttt{\small tooluniverse-expression-data-retrieval} & \small Retrieve gene expression and omics datasets from ArrayExpress and BioStudies with gene disambiguation and quality assessment\\
\texttt{\small tooluniverse-gene-enrichment} & \small Gene-set enrichment analysis: GO (Biological Process, Molecular Function, Cellular Component), KEGG, Reactome pathway enrichment...\\
\texttt{\small tooluniverse-functional-genomics-screens} & \small Interpret hits from CRISPR-KO/CRISPRi/shRNA screens by integrating DepMap essentiality, gnomAD constraint scores, pathway context...\\
\texttt{\small tooluniverse-crispr-screen-analysis} & \small Analyze CRISPR-Cas9 genetic screens: MAGeCK gene-level scores, sgRNA count QC, replicate correlation, hit prioritization, and...\\
\texttt{\small tooluniverse-gene-regulatory-networks} & \small Gene regulatory network analysis: TF-target inference (JASPAR motifs, ChIP-seq), motif scanning, eQTL integration, perturbation...\\
\texttt{\small tooluniverse-noncoding-rna} & \small Non-coding RNA analysis: miRNAs (miRBase, miRDB targets), lncRNAs (LNCipedia, RNAcentral), circRNAs, snoRNAs, and other ncRNA classes\\
\texttt{\small tooluniverse-epigenomics} & \small Genomics and epigenomics analysis: DNA methylation (CpG, 5mC, 5hmC, bisulfite, RRBS), m6A RNA modification (MeRIP-seq), ChIP-seq...\\
\texttt{\small tooluniverse-epigenomics-chromatin} & \small Histone-modification ChIP-seq, ATAC-seq accessibility, chromatin state, and TF binding analysis from ENCODE, Roadmap Epigenomics,...\\
\texttt{\small tooluniverse-regulatory-genomics} & \small Transcription factor binding, cis-regulatory elements (cCREs), chromatin accessibility, and regulatory annotation using JASPAR...\\
\midrule
\multicolumn{2}{l}{\cellcolor{gray!15}\textbf{Proteomics, protein structure and design}}\\ \midrule
\texttt{\small tooluniverse-proteomics-analysis} & \small Mass-spec proteomics analysis: protein identification, quantification (LFQ, TMT, iTRAQ), differential expression (tumor vs normal,...\\
\texttt{\small tooluniverse-proteomics-data-retrieval} & \small Find and retrieve proteomics datasets from MassIVE and ProteomeXchange\\
\texttt{\small tooluniverse-protein-interactions} & \small Protein-protein interaction (PPI) network analysis: STRING (predicted + experimental), BioGRID (curated), SASBDB (small-angle...\\
\texttt{\small tooluniverse-protein-structure-prediction} & \small Protein 3D structure prediction from sequence: ESMFold de novo prediction, AlphaFold database retrieval, experimental structures...\\
\texttt{\small tooluniverse-protein-structure-retrieval} & \small Protein structure retrieval from RCSB PDB, PDBe, and AlphaFold with disambiguation, quality assessment (resolution, R-factor,...\\
\texttt{\small tooluniverse-protein-therapeutic-design} & \small AI-guided de novo protein design: RFdiffusion backbone generation, ProteinMPNN sequence design, structure validation (pLDDT, pTM,...\\
\texttt{\small tooluniverse-protein-modification-analysis} & \small Post-translational modification (PTM) analysis: phosphorylation, ubiquitination, acetylation, glycosylation, methylation\\
\texttt{\small tooluniverse-protein-lof-mechanism} & \small Propose the mechanism by which a missense variant causes loss-of-function (LoF), synthesizing evidence from 5 independent layers:...\\
\texttt{\small tooluniverse-protein-sae-variant-interpretation} & \small Interpret a missense variant via ESMC-6B Sparse Autoencoder (SAE) feature activations\\
\texttt{\small tooluniverse-protein-structural-annotation-pdb} & \small Given a PDB structure, produce a per-residue annotation table: which residues sit at a binding interface (vs a partner chain),...\\
\texttt{\small tooluniverse-residue-functional-mechanism-interpretation} & \small Given a set of residues in a protein, explain WHY they are functionally critical by combining structural context (binding...\\
\texttt{\small tooluniverse-structural-proteomics} & \small Structural biology plus proteomics integration for drug target validation\\
\texttt{\small tooluniverse-variant-predictor-dms-validation} & \small Validate a variant-effect predictor (AlphaMissense, ESM-C SAE, ESM logits, EVE, conservation scores, or any per-variant numeric...\\
\texttt{\small tooluniverse-peptide-target-deorphanization} & \small Find the real protein target(s) of a peptide from its sequence: peptide target deorphanization / off-target identification, for...\\
\texttt{\small tooluniverse-electron-microscopy} & \small Search and analyze electron microscopy data: cryo-EM density maps (EMDB), fitted atomic models (PDB), raw micrograph datasets...\\
\texttt{\small tooluniverse-computational-biophysics} & \small Solve quantitative problems in biophysics: pharmacokinetics (PK volume of distribution, clearance, half-life), epidemiology (R0,...\\
\midrule
\multicolumn{2}{l}{\cellcolor{gray!15}\textbf{Metabolomics, lipidomics and enzymology}}\\ \midrule
\texttt{\small tooluniverse-metabolomics} & \small Metabolomics research: metabolite identification, study analysis, and database searches across HMDB, MetaboLights, Metabolomics...\\
\texttt{\small tooluniverse-metabolomics-analysis} & \small Analyze metabolomics data end-to-end: metabolite identification, quantification (TIC normalization, batch correction),...\\
\texttt{\small tooluniverse-metabolomics-pathway} & \small Metabolomics pathway analysis: metabolite identification (HMDB, KEGG, ChEBI), pathway mapping (Reactome, KEGG, MetaCyc), disease...\\
\texttt{\small tooluniverse-lipidomics} & \small Lipid analysis and lipid-disease associations using LIPID MAPS classification, HMDB metabolite data, KEGG/Reactome lipid pathways...\\
\texttt{\small tooluniverse-natural-product-dereplication} & \small Dereplicate a putative natural product and assign its chemical taxonomy\\
\texttt{\small tooluniverse-enzyme-kinetics} & \small Enzyme kinetics: Michaelis-Menten Km, Vmax, kcat (turnover), and kcat/Km (catalytic efficiency / specificity constant) from...\\
\midrule
\multicolumn{2}{l}{\cellcolor{gray!15}\textbf{Chemistry and small-molecule discovery}}\\ \midrule
\texttt{\small tooluniverse-admet-prediction} & \small Comprehensive ADMET (Absorption, Distribution, Metabolism, Excretion, Toxicity) profiling for drug candidates\\
\texttt{\small tooluniverse-binder-discovery} & \small Discover novel small-molecule binders for protein targets using structure-based and ligand-based screening\\
\texttt{\small tooluniverse-chemical-compound-retrieval} & \small Retrieve chemical compound data from PubChem and ChEMBL with disambiguation, cross-referencing, and stereochemistry handling\\
\texttt{\small tooluniverse-chemical-safety} & \small Chemical safety and toxicology assessment integrating ADMET-AI predictions, CTD toxicogenomics, PubChemTox experimental data,...\\
\texttt{\small tooluniverse-chemical-sourcing} & \small Find commercial sources for chemical compounds: PubChem/ChEMBL identity resolution then vendor catalog search across ZINC,...\\
\texttt{\small tooluniverse-small-molecule-discovery} & \small Small molecule identification, characterization, and procurement: PubChem, ChEMBL, BindingDB, ADMET-AI, SwissADME, eMolecules, Enamine\\
\texttt{\small tooluniverse-organic-chemistry} & \small Organic chemistry reasoning guide for reaction product prediction, mechanism analysis (electrophilic/nucleophilic substitution,...\\
\texttt{\small tooluniverse-inorganic-physical-chemistry} & \small Inorganic chemistry, physical chemistry, and materials science: crystal structures, coordination chemistry, lattice parameters,...\\
\texttt{\small tooluniverse-gpcr-structural-pharmacology} & \small GPCR receptor pharmacology: agonist/antagonist/inverse-agonist/biased-agonist classification, GPCRdb structural data,...\\
\texttt{\small tooluniverse-network-pharmacology} & \small Compound-target-disease network construction and analysis for drug repurposing, polypharmacology discovery, and multi-target drug...\\
\texttt{\small tooluniverse-drug-synergy} & \small Drug-combination synergy analysis: quantify whether two drugs together are synergistic, additive, or antagonistic using the...\\
\texttt{\small tooluniverse-dose-response} & \small Dose-response / concentration-response curve fitting: IC50, EC50, Hill slope, Emax/Emin efficacy, and relative potency from paired...\\
\midrule
\multicolumn{2}{l}{\cellcolor{gray!15}\textbf{Drug development and pharmacology}}\\ \midrule
\texttt{\small tooluniverse-drug-research} & \small Comprehensive drug profiling: mechanism, primary/secondary targets, drug interactions, clinical-trial status, adverse events...\\
\texttt{\small tooluniverse-drug-mechanism-research} & \small Trace drug mechanism of action: primary target  to  downstream signaling  to  pathway perturbation  to  tissue/organ effect  to  clinical outcome\\
\texttt{\small tooluniverse-drug-target-validation} & \small Quantitative drug-target validation pipeline\\
\texttt{\small tooluniverse-drug-repurposing} & \small Identify drug repurposing candidates via target-based, compound-based, and disease-based strategies\\
\texttt{\small tooluniverse-drug-drug-interaction} & \small Assess drug-drug interactions: CYP metabolic interactions (substrate/inhibitor/inducer), transporter (P-gp, BCRP, OATP) effects,...\\
\texttt{\small tooluniverse-drug-regulatory} & \small Drug regulatory and approval research: FDA substance registry, ATC/EPC classification, EMA decisions, generic-drug status, FDA...\\
\texttt{\small tooluniverse-target-research} & \small Comprehensive drug-target intelligence: tissue expression (GTEx, HPA), pathways, protein interactions (STRING), variant landscape...\\
\texttt{\small tooluniverse-pharmacokinetics} & \small Pharmacokinetic (PK) analysis of concentration-time data: non-compartmental analysis (NCA) for Cmax, Tmax, AUC (0-t and 0-infinity),...\\
\texttt{\small tooluniverse-pharmacogenomics} & \small Pharmacogenomics (PGx) research: drug-gene interactions (CPIC, PharmGKB), CPIC dosing guidelines, variant-drug-response...\\
\texttt{\small tooluniverse-antibody-engineering} & \small Therapeutic antibody engineering and optimization, lead-to-clinical-candidate\\
\texttt{\small tooluniverse-toxicology} & \small Drug and chemical toxicity assessment via adverse outcome pathways (AOPs), real-world FAERS adverse event signals, FDA labels, and...\\
\texttt{\small tooluniverse-adverse-outcome-pathway} & \small Map environmental and industrial chemicals to adverse outcome pathways (AOPs): molecular initiating event to organ-level toxicity\\
\midrule
\multicolumn{2}{l}{\cellcolor{gray!15}\textbf{Clinical, translational and safety}}\\ \midrule
\texttt{\small tooluniverse-clinical-data-integration} & \small End-to-end drug safety review integrating FDA labels, FAERS adverse event reports, PRR/ROR disproportionality, pharmacogenomic...\\
\texttt{\small tooluniverse-clinical-guidelines} & \small Search and retrieve clinical practice guidelines from 12+ authoritative sources: NICE, WHO, NCCN, AHA, ADA, SIGN, USPSTF, IDSA,...\\
\texttt{\small tooluniverse-clinical-risk-scoring} & \small Compute and interpret validated bedside clinical risk scores and pretest probabilities for an INDIVIDUAL patient: pick the right...\\
\texttt{\small tooluniverse-clinical-trial-design} & \small Strategic clinical trial design feasibility assessment\\
\texttt{\small tooluniverse-clinical-trial-matching} & \small AI-driven patient-to-trial matching for precision oncology and rare-disease care\\
\texttt{\small tooluniverse-precision-medicine-stratification} & \small Patient stratification for precision medicine: integrate genomic, clinical, and therapeutic data to split patients into...\\
\texttt{\small tooluniverse-precision-oncology} & \small Cancer treatment recommendations from molecular profile (mutations + cancer type + biomarkers): FDA-approved + investigational...\\
\texttt{\small tooluniverse-cancer-classification} & \small Translate free-text tumor descriptions to OncoTree codes and resolve cancer subtypes/tissue hierarchy\\
\texttt{\small tooluniverse-cell-line-profiling} & \small Cancer cell-line selection and profiling for experimental model choice\\
\texttt{\small tooluniverse-rare-disease-diagnosis} & \small Rare disease differential diagnosis from patient phenotype: HPO term matching to candidate diseases (Orphanet, OMIM), gene panel...\\
\texttt{\small tooluniverse-diagnostic-test-evaluation} & \small Diagnostic test / biomarker accuracy: sensitivity, specificity, PPV, NPV, likelihood ratios, accuracy from a 2x2 table; ROC curve,...\\
\texttt{\small tooluniverse-pharmacovigilance} & \small Drug safety and adverse event analysis: FAERS spontaneous-report mining, FDA black-box warnings, signal detection (PRR, ROR, IC),...\\
\texttt{\small tooluniverse-adverse-event-detection} & \small Detect and analyze adverse drug event signals using FDA FAERS reports, drug labels, and disproportionality statistics (PRR, ROR, IC)\\
\texttt{\small tooluniverse-product-safety-surveillance} & \small -\\
\texttt{\small tooluniverse-immunotherapy-response-prediction} & \small Predict patient response to immune checkpoint inhibitors (ICIs) by integrating tumor mutational burden (TMB), microsatellite...\\
\midrule
\multicolumn{2}{l}{\cellcolor{gray!15}\textbf{Immunology and infectious disease}}\\ \midrule
\texttt{\small tooluniverse-immunology} & \small Immunology research workflows: antibody-antigen interactions, T/B cell repertoire, MHC/HLA binding prediction, autoimmune disease...\\
\texttt{\small tooluniverse-immune-repertoire-analysis} & \small TCR/BCR repertoire analysis: V(D)J segment usage, CDR3 sequence diversity, clonality scoring, antigen specificity matching to...\\
\texttt{\small tooluniverse-hla-immunogenomics} & \small HLA gene-family analysis and MHC-peptide binding for transplant compatibility, vaccine epitope coverage, and cancer immunotherapy\\
\texttt{\small tooluniverse-vaccine-design} & \small Computational vaccine candidate design: peptide/subunit vaccines via MHC-I/MHC-II epitope prediction (IEDB), population HLA...\\
\texttt{\small tooluniverse-infectious-disease} & \small Rapid pathogen characterization and drug repurposing for outbreaks\\
\texttt{\small tooluniverse-microbiome-research} & \small Microbiome research using MGnify, GTDB, ENA, OLS (ENVO biomes), and EuropePMC\\
\texttt{\small tooluniverse-metagenomics-analysis} & \small Microbiome and metagenomics analysis using MGnify, GTDB taxonomy, ENA sequencing data, and EuropePMC literature\\
\midrule
\multicolumn{2}{l}{\cellcolor{gray!15}\textbf{Multi-omics, systems and disease biology}}\\ \midrule
\texttt{\small tooluniverse-multi-omics-integration} & \small Multi-omics integration: orchestrate per-layer analysis (transcriptomics, proteomics, epigenomics, genomics, metabolomics) then...\\
\texttt{\small tooluniverse-multiomic-disease-characterization} & \small Comprehensive disease characterization across genomics, transcriptomics, proteomics, and pathways for systems-level understanding\\
\texttt{\small tooluniverse-spatial-omics-analysis} & \small Spatial multi-omics interpretation pipeline\\
\texttt{\small tooluniverse-systems-biology} & \small Systems biology and pathway analysis integrating Reactome, KEGG, WikiPathways, BioCarta, NCI-Nature Pathway Interaction Database\\
\texttt{\small tooluniverse-disease-research} & \small Generate comprehensive disease research reports covering genetics (causal genes, GWAS, OMIM), pathways (Reactome, KEGG), drugs...\\
\texttt{\small tooluniverse-kegg-disease-drug} & \small KEGG-based disease-drug-variant network research\\
\texttt{\small tooluniverse-data-integration-analysis} & \small Integrate computed statistical results (DEGs, GWAS hits, associations) with biological context from ToolUniverse databases...\\
\texttt{\small tooluniverse-stem-cell-organoid} & \small Stem cell, iPSC, and organoid research: pluripotency markers, differentiation protocol pathways, lineage commitment factors,...\\
\texttt{\small tooluniverse-aging-senescence} & \small Aging biology, cellular senescence, and longevity research\\
\texttt{\small tooluniverse-neuroscience} & \small Neuroscience research workflows: neuroanatomy, neural circuits, neurotransmitter biology, neurological/psychiatric disease...\\
\midrule
\multicolumn{2}{l}{\cellcolor{gray!15}\textbf{Cross-cutting methods, data and literature}}\\ \midrule
\texttt{\small tooluniverse-literature-deep-research} & \small Deep literature review: PubMed, EuropePMC, bioRxiv preprints, citation networks, evidence synthesis\\
\texttt{\small tooluniverse-meta-analysis} & \small Meta-analysis / evidence synthesis: pool effect sizes across studies (odds ratios, risk ratios, hazard ratios, mean differences,...\\
\texttt{\small tooluniverse-statistical-modeling} & \small Statistical modeling: linear/logistic/ordinal/Poisson regression, ANOVA, Kruskal-Wallis, chi-square, Mann-Whitney, Cox survival,...\\
\texttt{\small tooluniverse-epidemiological-analysis} & \small End-to-end observational epidemiology analysis: from research question (PECO Population/Exposure/Comparator/Outcome) to...\\
\texttt{\small tooluniverse-dataset-discovery} & \small Find and evaluate research datasets for any scientific question\\
\texttt{\small tooluniverse-data-wrangling} & \small Universal data access patterns for downloading and parsing scientific data when ToolUniverse tools don't cover the source, only...\\
\texttt{\small tooluniverse-image-analysis} & \small Microscopy and quantitative imaging analysis: colony morphometry, fluorescence intensity quantification, cell-count statistics,...\\
\texttt{\small tooluniverse-phylogenetics} & \small Phylogenetic analysis: de novo multiple sequence alignment (Clustal Omega/MUSCLE/MAFFT via EBI\_msa\_align) and...\\
\texttt{\small tooluniverse-ecology-biodiversity} & \small Ecology, biodiversity, and conservation biology research: species identification (GBIF, NCBI Taxonomy), invasive species impact,...\\
\midrule
\bottomrule
\caption{The 130 research skills in \toun, grouped by application and data-modality area (as of 7 July 2026). Each skill is a natural-language-invokable workflow that orchestrates multiple tools into an end-to-end analysis; the skill catalogue is continually expanded.} \label{table:skill_categories}
\end{longtable}
\endgroup
\normalsize

}

\end{document}